\documentclass[10pt,journal,cspaper,compsoc]{IEEEtran}

\usepackage{algorithm}
\usepackage{booktabs}
\usepackage{epsfig}
\usepackage{graphicx}
\usepackage{amsmath}
\usepackage{amssymb}
\usepackage{color} 
\usepackage{algorithmic}
\usepackage{subfigure}
\usepackage{url}
\usepackage{multirow}
\usepackage[usenames,dvipsnames,table]{xcolor}
\usepackage[pagebackref=true,breaklinks=true,letterpaper=true,colorlinks,citecolor=citecolor,bookmarks=false]{hyperref}
\definecolor{citecolor}{RGB}{34,139,34}

\begin{document}

\title{Cross-Modal Progressive Comprehension for Referring Segmentation}

\author{Si Liu, Tianrui Hui, Shaofei Huang, Yunchao Wei, Bo Li, Guanbin Li\thanks{
Si Liu and Bo Li are with Institute of Artificial Intelligence, Beihang University, Email: \{liusi, boli\}@buaa.edu.cn.
Tianrui Hui and Shaofei Huang are with Institute of Information Engineering, Chinese Academy of Sciences, and also with School of Cyber Security, University of Chinese Academy of Sciences, E-mail: \{huitianrui, huangshaofei\}@iie.ac.cn.
Yunchao Wei is with Institute of Information Science, Beijing Jiaotong University, Email: wychao1987@gmail.com.
Guanbin Li is with School of Computer Science and Engineering, Sun Yat-sen University, and also with Pazhou Lab, E-mail: liguanbin@mail.sysu.edu.cn.
Corresponding author: Guanbin Li.}}

\markboth{IEEE TRANSACTIONS ON PATTERN ANALYSIS AND MACHINE INTELLIGENCE, VOL. XX, NO. X, X 20XX}
{Shell \MakeLowercase{\textit{et al.}}: Bare Demo of IEEEtran.cls for Computer Society Journals}

\IEEEcompsoctitleabstractindextext{
\begin{abstract} 
   Given a natural language expression and an image/video, the goal of referring segmentation is to produce the pixel-level masks of the entities described by the subject of the expression.
   Previous approaches tackle this problem by implicit feature interaction and fusion between visual and linguistic modalities in a one-stage manner. However, human tends to solve the referring problem in a progressive manner based on informative words in the expression, i.e., first roughly locating candidate entities and then distinguishing the target one.
   In this paper, we propose a Cross-Modal Progressive Comprehension (CMPC) scheme to effectively mimic human behaviors and implement it as a CMPC-I (Image) module and a CMPC-V (Video) module to improve referring image and video segmentation models. 
   For image data, our CMPC-I module first employs entity and attribute words to perceive all the related entities that might be considered by the expression. 
   Then, the relational words are adopted to highlight the target entity as well as suppress other irrelevant ones by spatial graph reasoning. 
   For video data, our CMPC-V module further exploits action words based on CMPC-I to highlight the correct entity matched with the action cues by temporal graph reasoning. 
   In addition to the CMPC, we also introduce a simple yet effective Text-Guided 
   Feature Exchange (TGFE) module to integrate the reasoned multimodal features corresponding to different levels in the visual backbone under the guidance of textual information. 
   In this way, multi-level features can communicate with each other and be mutually refined based on the textual context. 
   Combining CMPC-I or CMPC-V with TGFE can form our image or video version referring segmentation frameworks and our frameworks achieve new state-of-the-art performances on four referring image segmentation benchmarks and three referring video segmentation benchmarks respectively.
   Our code is available at \url{https://github.com/spyflying/CMPC-Refseg}.
\end{abstract}

\begin{keywords}
Referring Segmentation, Progressive Comprehension, Graph Reasoning, Multimodal Feature Fusion.
\end{keywords}
}

\maketitle

\IEEEdisplaynotcompsoctitleabstractindextext

\IEEEpeerreviewmaketitle

\section{Introduction}

In this paper, we target at an emerging task called \textit{referring segmentation}~\cite{hu2016segmentation}\cite{li2018referring}\cite{chen2019see}. 
Given a natural language expression and an image/video as inputs, the goal of referring segmentation is to segment the entities referred by the subject of the input expression. 
Traditional semantic segmentation methods~\cite{chen2017deeplab}\cite{zhao2017pyramid}\cite{fu2019dual} aim to classify each pixel as one of a fixed set of categories denoted by short words (e.g., ``person'', ``cell phone''). 
Referring segmentation can be regarded as a generalized semantic segmentation task where the categories belong to an open set denoted by expressions with various grammars and diverse contents, such as entities, attributes, relationships and actions, etc. 
Combining rich visual and linguistic information, referring segmentation has a wide range of potential applications such as language-based robot controlling~\cite{wang2019reinforced}\cite{gao2021room}, interactive image editing~\cite{chen2018language}, etc.

Previous works tackle the referring image or video segmentation task using a straightforward concatenation-and-convolution scheme~\cite{hu2016segmentation} involving dynamic filters~\cite{margffoy2018dynamic}\cite{gavrilyuk2018actor}, convolutional RNNs~\cite{liu2017recurrent}\cite{li2018referring} or cross-modal attention mechanism~\cite{shi2018key}\cite{chen2019see}\cite{ye2019cross}\cite{wang2019asymmetric} to fuse visual and linguistic features. 
Instead of the above one-stage implicit approaches, human tends to comprehend the referring expression in a progressive way~\cite{tanenhaus1995integration}. 
As human reads the expression, different nouns and adjectives will first be located in the image/video to find all the candidate entities. Then, relational and action words like prepositions and verbs are extracted to reason the relationships among different entities, where the disturbing entities are excluded and the target one is found out.

\begin{figure*}[t]
   \begin{center}
      \includegraphics[width=1.0\linewidth]{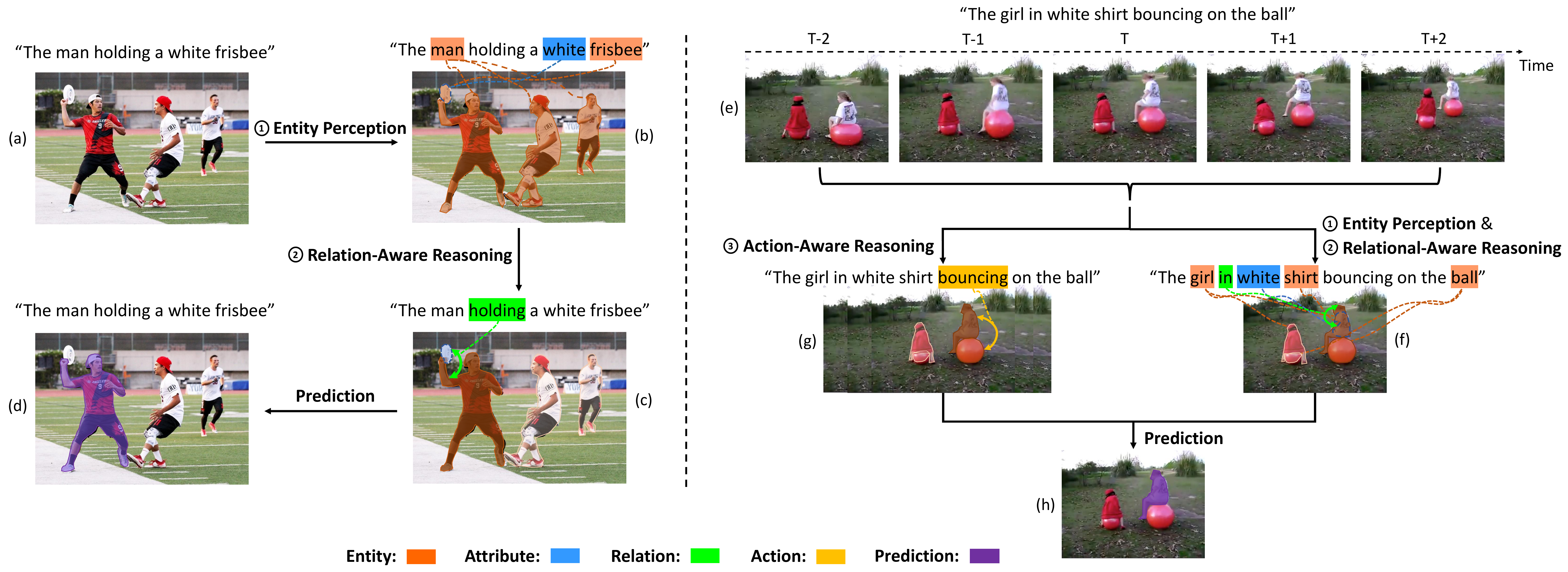}
   \end{center}
      \caption{Illustration of our progressive referring segmentation method on image and video data. 
      (a)(e) Input referring expression, image and video. 
      (b)(f) Our model first perceives all the entities 
      described in the expression based on entity words and attribute words, e.g., ``man'', ``white frisbee'' and ``girl'' (orange masks and blue outline). 
      (c)(f) After finding out all the candidate entities that may be referred by the input expression, our model exploits relational word, e.g., ``holding'' and ``in'', to highlight the entity involved with the relationship (green arrow) and suppress the others which are not involved. 
      (g) For video data, our model further utilizes action word ``bouncing'' to aggregate temporal context information from neighbor frames for better localizing the referent in a changing video. 
      (d)(h) Based on the progressive comprehension results, our model can finally determine the referent as output (purple mask). (Best viewed in color).}
   \label{fig:intro}
\end{figure*}

To mimic the more natural processing way of human beings, we introduce a Cross-Modal Progressive Comprehension (CMPC) scheme to solve the referring segmentation task in multiple stages. 
We define the entity referred by the expression as the \textit{referent}. 
For referring image segmentation, we illustrate the CMPC scheme in the left part of Fig.~\ref{fig:intro}. 
If the referent is described by ``The man holding a white frisbee'', the referring process is divided into two progressive stages. 
First, the model can utilize entity words and attribute words, e.g., ``man'' and ``white frisbee'', to perceive all the possible entities mentioned in the expression. Second, as one image may contain multiple entities of the same category, for example, the three men in Fig.~\ref{fig:intro} (b), the model needs to further highlight the referent matched with the relationship cue in the expression while suppressing other mismatched ones by reasoning relationships among entities. 
In Fig.~\ref{fig:intro} (c), under the guidance of the relational word ``holding'' which associates ``man'' with ``white frisbee'', the model can focus on the referent who holds a white frisbee rather than the other two men.
After comprehending multimodal information progressively, the model can make correct prediction as shown in Fig.~\ref{fig:intro} (d).

Different from image, expressions containing action descriptions~\cite{gavrilyuk2018actor}\cite{wang2019asymmetric} are often used to refer entities in a video. 
In the right part of Fig.~\ref{fig:intro}, we illustrate the video version of our CMPC scheme.
If the entity in a video is described by ``The girl in white shirt bouncing on the ball'', the referring process is divided into three stages. 
The first entity perception stage and the second relation-aware reasoning stage are conducted on the center frame of a video snippet, which is almost same as those of the image model.
Then in the third stage, action words (e.g., ``bouncing'') are exploited by the model to capture temporal cues among frames in the video to further highlight the referent conducting the described action, as shown in Fig.~\ref{fig:intro} (g).
Finally, multimodal spatio-temporal information is comprehended by the model to make correct predictions along the video as shown in Fig.~\ref{fig:intro} (h).

To tackle image and video inputs respectively, we develop two versions of CMPC scheme. 
\textbf{For image data}, we propose a CMPC-I (Image) module which progressively exploits different types of words in the expression to segment the referent in an image. 
Concretely, our CMPC-I module consists of two stages. 
First, we extract linguistic features of entity words and attribute words 
(e.g., ``man'' and ``white frisbee'') from the expression and then fuse them with visual features extracted from the image to build multimodal features. 
During this process, all the entities that may be referred by the expression are perceived. 
Second, we construct a fully-connected spatial graph where each image region is regarded as a vertex and multimodal information of the entity is contained in each vertex. 
Appropriate edges are required for vertexes to communicate with each other. 
Naive edges which connect all the vertexes equally will introduce abundant information and hinder the identification of the referent. 
Thus, our CMPC-I module employs relational words (e.g., ``holding'') of the expression as a group of routers to build adaptive edges to connect spatial vertexes, i.e., entities, which are involved with the relationship described in the expression. 
Particularly, spatial vertexes (e.g., ``man'') yielding strong responses to the relational words (e.g., ``holding'') will exchange information with other vertexes (e.g., ``frisbee'') that also highly correlate with the relational words. 
Meanwhile, less interaction will occur among spatial vertexes yielding weak responses to the relational words. 
After relation-aware reasoning on the multimodal graph, our CMPC-I module can highlight feature of the referent while suppressing those of the irrelevant entities, which assists in generating accurate segmentation.
\textbf{For video data}, we further extend our CMPC-I module to CMPC-V (Video) module with an additional action-aware reasoning stage based on action words to exploit temporal information. 
Concretely, our CMPC-V module extracts global multimodal features of all the frames in the video snippet based on action words (e.g., ``bouncing'') after the same entity perception and relation-aware reasoning stages.
A fully-connected temporal graph is constructed using each frame as a vertex where frame feature is served as vertex feature. 
Information propagation among temporal vertexes is performed to extract temporal multimodal context among frames. 
Finally, features of the temporal graph are aggregated with feature of annotated frame to supplement temporal multimodal context for better segmentation.

As prior works~\cite{li2018referring}\cite{ye2019cross}\cite{chen2019see} show multiple levels of visual features can complement each other, we also propose a Text-Guided Feature Exchange (TGFE) module to exploit information of multimodal features refined by our CMPC modules from different levels. 
Combining CMPC-I or CMPC-V module with TGFE module forms our image or video version referring segmentation framework. 
For each level of multimodal features, our TGFE module utilizes linguistic features 
as guidance to select useful feature channels from other levels to enable information 
communication. 
After multiple rounds of communication, Our TGFE further fuses multi-level features 
by ConvLSTM~\cite{xingjian2015convolutional} to comprehensively integrate low-level visual details and high-level semantics for precise segmentation results.

The main contributions of our paper are summarized as follows:
\begin{itemize}
   \item We introduce a Cross-Modal Progressive Comprehension (CMPC) scheme to align multimodal features in multiple stages based on informative words in the expression, which provides a general solution to referring segmentation task and is robust to different visual modalities.
   \item We instantiate the CMPC scheme by proposing a CMPC-I module containing entity perception and relation-aware reasoning stages for referring image segmentation. We further propose a CMPC-V module by extending CMPC-I with an action-aware reasoning stage for referring video segmentation.
   \item We also propose a TGFE module to aggregate multi-level multimodal features to enhance feature of the referent for better segmentation.
   \item Combining CMPC-I or CMPC-V with TGFE, our image and video version frameworks achieves current state-of-the-art performances on four referring image segmentation benchmarks and three referring video segmentation benchmarks, respectively.
\end{itemize}

This paper is an extension of our previous conference version~\cite{huang2020cmpc}. 
The current work adds to the initial version with significant aspects. 
First, we extend our CMPC scheme from referring image segmentation to referring video segmentation by introducing an additional action-aware reasoning stage, which effectively extracts temporal multimodal context to enhance feature of the referent. 
Second, we improve our initial CMPC-I module by changing the way of multimodal feature concatenation to better highlight feature of the referent. 
Third, we add considerable new experimental results including ablation study, model setting and visualization analysis. 
Our image model in this paper also obtains better performance than our conference version.

\section{Related Work}
\subsection{Semantic Segmentation}
Based on Fully Convolutional Networks (FCN)~\cite{long2015fully}, semantic segmentation has made a huge progress in recent years. 
FCN uses convolution layers to replace all the fully-connected layers in original classification networks and becomes the most popular architecture in the semantic segmentation community. 
DeepLab series~\cite{chen2014semantic}\cite{chen2017deeplab}\cite{chen2017rethinking} 
incorporates FCN with dilated convolutions with different dilation rates, which enlarges the receptive field of filters to aggregate multi-scale visual context. 
PSPNet~\cite{zhao2017pyramid} proposes similar pyramid pooling operations to extract 
multi-scale context as well. Later works such as DANet~\cite{fu2019dual} and 
CFNet~\cite{zhang2019co} exploit self-attention mechanism~\cite{wang2018non} to 
capture long-range dependencies among image positions and achieve notable performance. 
In this paper, we target at the more challenging semantic segmentation problem whose semantic categories are specified by diverse natural language sentences.

\subsection{Referring Expression Grounding}
Given a natural language expression, referring expression grounding aims to localize the entities matched with the expression in the given image or video. 
Many works conduct localization in bounding box level. 
Liao \textit{et al.}~\cite{liao2019real} performs cross-modality correlation filtering to match features from different modalities in real time. 
Graph models involving attention mechanism are explored in~\cite{yang2019cross}\cite{yang2019dynamic}\cite{yang2020graph}\cite{yang2020relationship} to find the most related objects for the expression. 
Yu \textit{et al.}~\cite{yu2018mattnet} propose modular networks to decompose the referring expression into subject, location and relationship in order to finely compute the matching score. 
Most box-based methods are two-stage where a pretrained detector is utilized to first generate RoI proposals for later grounding. This design paradigm achieves high localization performance but lacks global context information and heavily relies on the quality of proposal candidates. 
In addition, it can only ground a single object and cannot locate the stuff or multiple objects.

Beyond bounding box, the referred object can also be localized more precisely with segmentation mask. 
Hu \textit{et al.}~\cite{hu2016segmentation} first proposes the referring image segmentation problem and directly concatenates and fuses multimodal features from CNN and 
LSTM~\cite{hochreiter1997long} to generate the segmentation mask. 
In~\cite{liu2017recurrent} and~\cite{margffoy2018dynamic}, multimodal LSTM is employed to 
sequentially fuse visual and linguistic features in multiple time steps. 
Multi-level feature fusion is explored in~\cite{li2018referring} to recurrently refine the local details of segmentation mask. 
As context information is critical to segmentation task, recent works employ cross-modal attention~\cite{shi2018key}\cite{chen2019see}\cite{Ye2020DualCL} and self-attention~\cite{ye2019cross} to extract multimodal context between image regions and referring words. 
Cycle-consistency learning~\cite{chen2019referring} and adversarial training~\cite{qiu2019referring} are also investigated to boost the segmentation performance. 
Gavrilyuk \textit{et al.}~\cite{gavrilyuk2018actor} further introduce referring segmentation task into video data in which sentences contain action descriptions of the actors in the videos. 
They utilize dynamic filters and multi-resolution decoder to generate mask of the referent. 
Based on~\cite{gavrilyuk2018actor}, Wang \textit{et al.}~\cite{wang2019asymmetric} propose asymmetric cross-guided attention between visual and linguistic modalities to segment the referent more precisely. 
Different from box-based methods, most mask-based methods are one-stage where FCN~\cite{long2015fully} is utilized to directly generate the mask of the referent. 
This design paradigm can extract rich global context information and does not rely on pretrained detectors. 
However, generating masks based on monolithic representations of referring expressions and visual contents may be difficult to distinguish between different instances, thus producing false positive masks and harming localization performance. 
In this paper, we tackle the above issue of one-stage methods by proposing to progressively highlight the referent via entity perception, relation-aware reasoning and action-aware reasoning, which effectively distinguishes different instances and reasons the referent on both image and video data.

\subsection{Graph-Based Reasoning}
Recently, graph-based models have shown their effectiveness in context reasoning for many tasks. 
Graph Convolution Networks (GCN)~\cite{chandra2017dense} becomes popular for its superiority on semi-supervised classification. 
Wang \textit{et al.}~\cite{wang2018videos} uses RoI proposals as vertexes to construct a spatial-temporal graph and conduct context reasoning with GCN, which improve performance on video recognition task. 
Chen \textit{et al.}~\cite{chen2019graph} propose 
a global reasoning module which projects visual feature into an interactive space and performs graph convolution for global context reasoning. 
Then they reproject the reasoned global context back to the coordinate space 
to enhance original visual feature. 
There are several concurrent works~\cite{li2018beyond}\cite{liang2018symbolic}\cite{zhang2019latentgnn} sharing the same spirit with~\cite{chen2019graph} while using different implementations. 
Graph-based reasoning has also been widely applied in vision and language tasks such as Hu~\textit{et al.}~\cite{hu2019language} and Yu~\textit{et al.}~\cite{yu2019heterogeneous}. Hu~\textit{et al.}~\cite{hu2019language} build a fully-connected visual graph where each node corresponds to an object proposal generated by a pre-trained detector and formulates the message passing among graph nodes as a recurrent process. Yu~\textit{et al.}~\cite{yu2019heterogeneous} build two heterogeneous graphs where the primal vision-to-answer graph utilizes object proposals and answer words as graph nodes, while the dual question-to-answer graph utilizes query words and answer words as graph nodes. 
They conduct message passing between visual nodes and answer nodes in the primal graph, while between query nodes and answer nodes in the dual graph.
Different from above two methods, we propose to regard image regions and video frames as vertexes to build 
spatial and temporal graphs for effectively reasoning multimodal context based on informative words in the expression which is more sutible for the segmentation task.
Besides, we exploit the relational words as routers to connect each pair of visual nodes on the feature map and message passing among all visual nodes is guided by relational words in a more effective way.

\begin{figure*}[t]
   \begin{center}
      \includegraphics[width=1.0\linewidth]{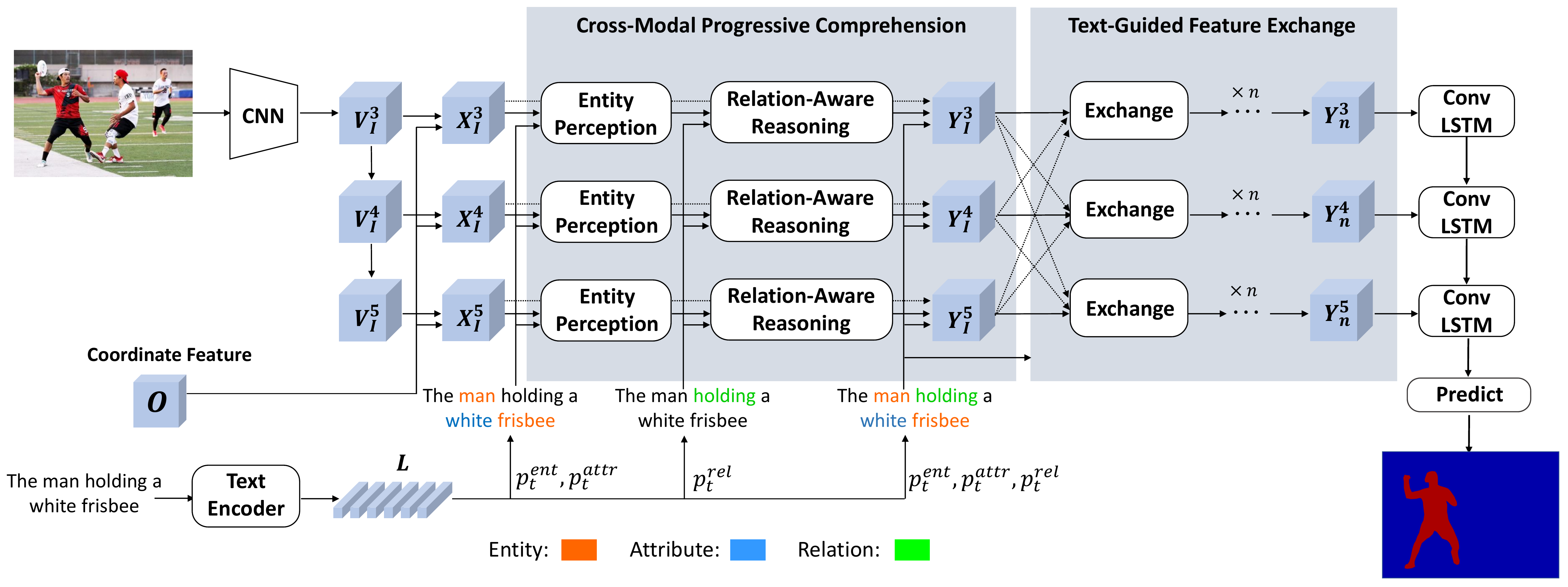}
   \end{center}
      \caption{Overview of our proposed method using referring image segmentation as an example. Visual features and linguistic features are first progressively aligned by 
      our CMPC-I module. Then multi-level multimodal features are fed into our TGFE module for information communication across different levels. Finally, multi-level features are fused with ConvLSTM for final prediction.}
   \label{fig:pipeline}
\end{figure*}

\section{Method}
In this section, we elaborate the instantiations of our introduced CMPC scheme to tackle the referring segmentation task on image and video data respectively. 
The proposed modules are denoted as CMPC-I (Image) and CMPC-V (Video) in the rest of our paper.

\subsection{CMPC on Image}
\subsubsection{Image and Words Feature Extraction}
As shown in Fig.~\ref{fig:pipeline}, our image model takes an image and a sentence as inputs. 
We use a CNN backbone to extract multi-level image features $V_I$ and fuse them with an $8$-D spatial coordinate feature $O \in \mathbb{R} ^ {H \times W \times 8}$ respectively using a $1 \times 1$ convolution layer following prior works~\cite{liu2017recurrent}\cite{ye2019cross}. 
After the convolution, each level of image features are transformed to the 
same size of $\mathbb{R} ^ {H \times W \times C_v}$, 
with $H$, $W$ and $C_v$ being the height, width and channel dimension of the image features, respectively. 
We denote the transformed image features by \{$X_I^3$, $X_I^4$, $X_I^5$\} which correspond to the output of the $3$rd, $4$th and $5$th stages of CNN backbone (e.g., ResNet-101~\cite{he2016deep}).
For ease of presentation, we denote a single level of image features as $X_I \in \mathbb{R}^{H \times W \times C_v}$. 
The words features $L = \{l_1, l_2, ..., l_T\}$ is extracted with a language 
encoder (e.g., LSTM~\cite{hochreiter1997long}), where $T$ is the length 
of the sentence and $l_i \in \mathbb{R} ^ {C_l} (i \in \{1, 2, ..., T\})$ denotes 
feature of the $i$-th word.

\subsubsection{Entity Perception}
\label{sec:ep}
Since the input image may contain many entities, it is natural to progressively narrow down the candidate set from all the entities to the actual referent. 
The first stage of our CMPC-I (Image) module is entity perception. 
We associate linguistic features of entity words and attribute words 
with the correlated visual features of spatial regions using bilinear fusion~\cite{BenMUTAN} to perceive all the candidate entities. 

Concretely, we classify the words into $4$ types, i.e., entity, attribute, relation and unnecessary word following~\cite{yang2019cross}. 
Since there is no annotation for word types of referring expressions in the datasets we used, we do not have direct supervision on learning the word classification. 
Thus, we supervise the word classification using the binary cross-entropy segmentation loss at the end of our model, which is also adopted by~\cite{yang2019cross}. 
Detailed experimental analyses and visualized results about word classification are presented in Section~\ref{sec:visualization}.
For each word, a $4$-D vector is predicted to indicate the probability that it belongs to one of these $4$ categories. 
We denote the probability vector of word $t$ as $p_t = [p_{t}^{ent}, p_{t}^{attr}, p_{t}^{rel}, p_{t}^{un}] \in \mathbb{R} ^ 4$ and calculate it by:
\begin{equation}
   \label{eq:prob_pred}
   p_t = softmax(W_2\sigma(W_1l_t+b_1)+b_2),
\end{equation}
where $W_1 \in \mathbb{R} ^ {C_n \times C_l}$, $W_2 \in \mathbb{R} ^ {4 \times C_n}$, $b_1 \in \mathbb{R} ^ {C_n}$ and $b_2 \in \mathbb{R} ^ {4}$ are learnable parameters, $\sigma(\cdot)$ is sigmoid function, 
$p_{t}^{ent}$, $p_{t}^{attr}$, $p_{t}^{rel}$ and $p_{t}^{un}$ denote the probabilities of word $t$ being the entity, attribute, relation and unnecessary word respectively. 
Then, we calculate the global linguistic context of entities $q \in \mathbb{R} ^ {C_l}$ by weighted combination of the all the words in the expression: 
\begin{equation} 
   \label{eq:setence_feat}
   q_e = \sum\limits_{t=1}^{T}{(p_t^{ent}+p_t^{attr})l_t}.
\end{equation}

Next, we adopt a simplified bilinear fusion strategy~\cite{BenMUTAN} to associate $q_e$ with the image feature $X_I$ on each spatial region to obtain the multimodal feature $M_I \in \mathbb{R}^{H \times W \times C_m}$ as follows:
\begin{equation}
   \label{eq:mutan}
   M_I^i = (q_eW_{3i})\odot (X_IW_{4i}),
\end{equation}
\begin{equation}
   \label{eq:ta}
   M_I = \sum\limits_{i=1}^{r}{M_I^i}
\end{equation}
where $W_{3i} \in \mathbb{R}^{C_l \times C_m}$ and $W_{4i} \in \mathbb{R}^{C_v \times C_m}$ are learnable parameters, $\odot$ denotes element-wise product and $r$ is a hyper-parameter.
By integrating both visual and linguistic context into the multimodal features, all the entities that may be matched with the expression are accordingly perceived.

\begin{figure*}[t]
   \begin{center}
      \includegraphics[width=0.9\linewidth]{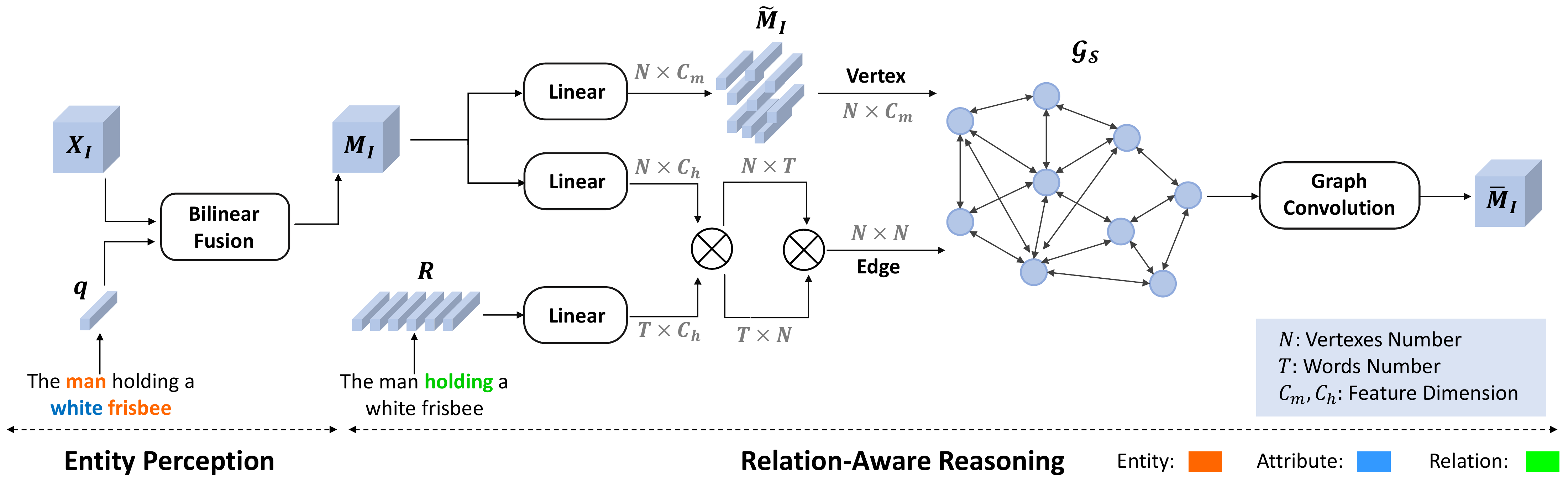}
   \end{center}
      \caption{Illustration of our CMPC-I module which consists of two stages. 
      First, visual features $X_I$ are bilinearly fused with linguistic features $q$ of entity words and attribute words for Entity Perception (EP) stage. 
      Second, multimodal features $M_I$ from EP stage are fed into Relation-Aware Reasoning (RAR) stage for feature enhancement. 
      A multimodal fully-connected graph $\mathcal{G_S}$ is constructed with each vertex corresponds to an image region on $M$. 
      The adjacency matrix of $\mathcal{G_S}$ is defined as the product of the matching degrees between vertexes and relational words in the expression. 
      Graph convolution is utilized to reason among vertexes so that the referent could be 
      highlighted during the interaction with correlated vertexes.}
   \label{fig:reason}
\end{figure*}

\subsubsection{Relation-Aware Reasoning}
\label{sec:rar}
After perceiving all the possible entities in the image, the second stage of our CMPC-I module is relation-aware reasoning. 
We construct a fully-connected multimodal graph based on $M_I$ using relational words as a group of routers to connect vertexes. 
Each vertex of the graph represents a spatial region on $M_I$. 
By reasoning among vertexes of the multimodal graph, our model can highlight the responses of the referent which are involved with the relationship cue while suppressing those of the non-referred ones. 

Concretely, the multimodal graph is defined as $\mathcal{G_S}  = (\mathcal{V}, \mathcal{E}, \tilde{M_I}, A)$ where $\mathcal{V}$ and $\mathcal{E}$ are the sets of vertexes and edges, $\tilde{M_I} = \{\tilde{m_I}^j\}_{j=1}^{N} \in \mathbb{R} ^ {N \times C_m}$ is the set of vertex features, $A \in \mathbb{R} ^ {N \times N}$ is the adjacency matrix and $N$ is number of vertexes. 

Details of relation-aware reasoning is illustrated in the right part of Fig.~\ref{fig:reason}. 
Since each location on $M_I$ represents a spatial region on the original image, each region is regarded as a vertex of the graph. 
The multimodal graph consists of $N = H \times W$ vertexes in total.
A linear layer is applied to $M_I$ to transform it into the features of vertexes $\tilde{M_I}$ after the reshaping operation. 
We use the affinities between vertexes and relational words in the referring expression to determine the edge weights. 
Features of relational words $R = \{r_t\}_{t=1}^T \in \mathbb{R} ^{T \times C_l}$ are calculated by:
\begin{equation}
   \label{eq:relation-extract}
   r_t = p_t^{rel}l_t,~~~~t=1, 2, ..., T.
\end{equation}

As shown in Fig.~\ref{fig:reason}, to obtain adjacency matrix $A$, we first conduct cross-modal attention mechanism by matrix product between $\tilde{M}_I$ and $R$ with necessary dimension transformation: 
\begin{equation}
   \label{eq:inter}
   B = (\tilde{M_I}W_5) (RW_6)^T,
\end{equation}
where $B \in \mathbb{R}^{N \times T}$ measures the feature relevance between each vertex of $\tilde{M_I}$ and each word of $R$. 
Then, we apply softmax function on the $T$ dimension of $B$ and obtain $B1 \in \mathbb{R}^{N \times T}$: 
\begin{equation}
   \label{eq:u1}
   B_1 = softmax(B),
\end{equation}
where each row of $B1$ means the normalized $T$ attention weights between each vertex and $T$ words. 
Accordingly, another softmax on the $N$ dimension of $B^{T} \in \mathbb{R}^{T \times N}$ obtains $B2 \in \mathbb{R}^{T \times N}$: 
\begin{equation}
   \label{eq:u2}
   B_2 = softmax(B^T),
\end{equation}
where each row of $B2$ means the normalized $N$ attention weights between each word and $N$ vertexes. The meaning of softmax operation is normalizing the attention weights to control the amount of information propagation on the graph within a reasonable interval.
Afterwards, we perform matrix product between $B1 \in \mathbb{R}^{N \times T}$ and $B2 \in \mathbb{R}^{T \times N}$ and obtain normalized adjacency matrix $A \in \mathbb{R}^{N \times N}$: 
\begin{equation}
   \label{eq:adj_matrx}
   A = B_1 B_2.
\end{equation}

Each element $A_{ij}$ of $A$ represents the normalized magnitude of information flow from the spatial region $i$ to the region $j$, which depends on their affinities with relational words in the expression. If a vertex has high attention weight with a certain word, and this word also has high attention weight with another vertex, then these two vertexes will have high attention weight with each other in a relational context.
In this way, adaptive edges connecting spatial vertexes can be built by leveraging relational words of the expression as a group of routers.

After building the multimodal graph $\mathcal{G_S}$, we conduct graph convolution~\cite{kipf2016semi} over it as follow:
\begin{equation}
   \label{eq:gcn}
   \bar{M_I} = (A+I)\tilde{M_I}W_7,
\end{equation}
where $W_7 \in \mathbb{R} ^{C_m \times C_m}$ is a learnable weight matrix, 
$I$ is the identity matrix serving as a residual connection to ease optimization.
The graph convolution performs reasoning among vertexes, i.e., image regions, which selectively highlights the referent according to the relationship cues while suppressing other irrelevant ones. As a result, more discriminative features can be generated to improve referring segmentation.

Afterwards, we conduct reshaping operation to obtain the image-format enhanced multimodal 
features $\bar{M_I} \in \mathbb{R} ^ {H \times W \times C_m}$.
To exploit the textual information, we first aggregate features of all necessary 
words into a vector $s \in \mathbb{R} ^ {C_l}$ by weighted sum using the predefined probability vectors as weights:
\begin{equation}
   \label{eq:sentence}
   s = \sum_{t=0}^{T}{(p_t^{ent}+p_t^{attr}+p_t^{rel})l_t}.
\end{equation}
Then, we repeat $s$ for $H \times W$ 
times and concatenate it with $M_I$ and $\bar{M_I}$ along channel dimension. 
In our conference version~\cite{huang2020cmpc}, we concatenate repeated $s$ with image feature $X_I$ and $\bar{M_I}$. 
Since the multimodal feature $M_I$ contains richer context information about entities in the image, we suppose concatenating $M_I$ can provide more sufficient guidance than pure image feature $X_I$, which is also proved empirically in Table~\ref{tab:sota_ris} and Table~\ref{tab:cmpc_i}. 
We further apply a $1 \times 1$ convolution on the concatenated features to obtain the output feature $Y_I \in \mathbb{R} ^ {H \times W \times C_m}$, which embodies multimodal context for the referent.

\begin{figure*}[t]
   \begin{center}
      \includegraphics[width=0.9\linewidth]{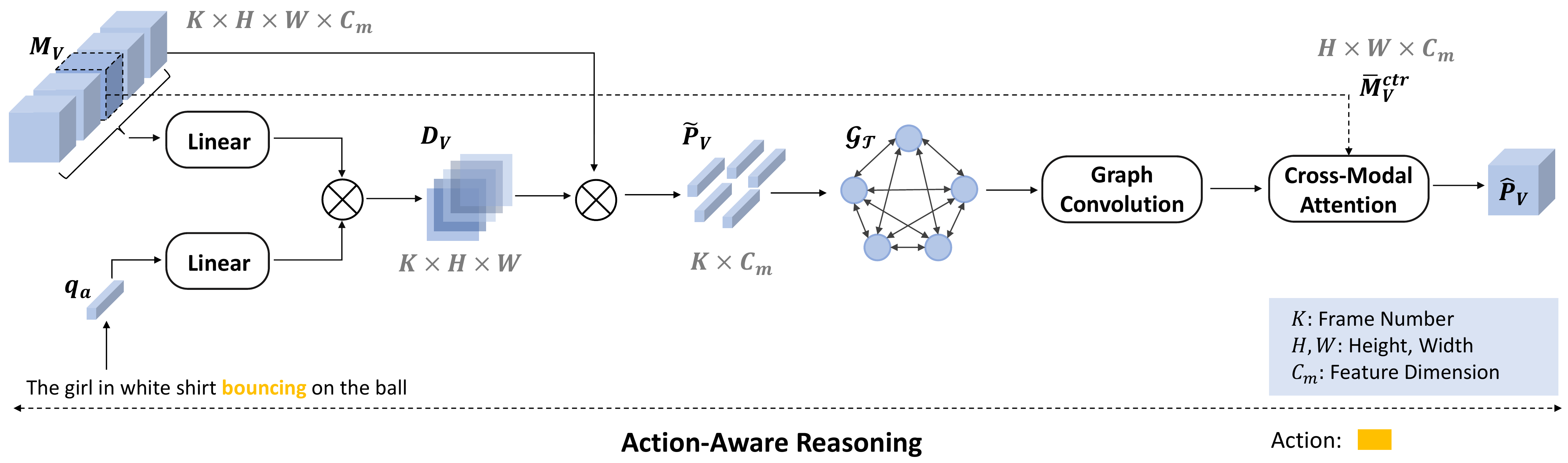}
   \end{center}
      \caption{Illustration of the action-aware reasoning stage of our CMPC-V module. 
      We ignore previous EP and RAR stages for clarity. 
      We first conduct dot-product attention between video features $M_V$ and action sentence feature $q_a$ to obtain the dense attention maps $D_V$ on all the frames of the video snippet. 
      Then, $D_V$ is applied on $M_V$ to aggregate global temporal features $\tilde{P}_V$ of all the frames. 
      We construct a fully-connected temporal graph $\mathcal{G_T}$ based on $\tilde{P_V}$ and perform graph convolution on it to reason temporal context. 
      Finally, reasoned temporal context are projected back to the feature of the center frame $\bar{M}_V^{ctr}$ to yield the image-format temporal context $\hat{P}_V$.}
   \label{fig:cmpcv}
\end{figure*}

\subsection{CMPC on Video}
\subsubsection{Video and Sentence Feature Extraction}
Our video model takes a video clip and a sentence as input. 
The length of the video clip is $K$ frames and the center frame is annotated with pixel-wise mask. 
We reshape the video clip as a batch of images and feed them into the CNN backbone as we do in the CMPC-I model. 
The output multi-level video clip features are denoted as $V_F$. 
We follow the same protocol in the image model to fuse each level of $V_F$ with $8$-D coordinate feature $O \in \mathbb{R} ^ {H \times W \times 8}$ respectively using a $1 \times 1$ convolution. 
After the convolution, each level of video clip features are transformed to the 
same size of $\mathbb{R}^{K \times H \times W \times C_f}$, 
with $K$, $H$, $W$ and $C_f$ being the frame numbers, height, width and channel dimension of the video clip features. 
We denote the transformed video clip features by \{$X_F^3$, $X_F^4$, $X_F^5$\} which correspond to the output of the $3$rd, $4$th and $5$th stages of CNN backbone (e.g., ResNet-101~\cite{he2016deep}) and denote a single level of video clip features as $X_F$ for ease of presentation. 
The extraction of words features $L = \{l_1, l_2, ..., l_T\} \in \mathbb{R}^{T \times C_l}$ is same as our image model.

\subsubsection{Action-Aware Reasoning}
As video data usually contains temporal information, reasoning only static relationship between entities are not enough to identify the referent in a video. 
Therefore, we further introduce an action-aware reasoning stage to highlight entities matched with the temporal action cues and fuse them with entities matched with the static relationship cues to distinguish the referent. 

Concretely, we first use a similar approach to the Section~\ref{sec:ep} to classify the words into $5$ types, i.e., entity, attribute, relation, action and unnecessary word. 
We denote the probability vector of word $t$ as $p_t = [p_{t}^{ent}, p_{t}^{attr}, p_{t}^{rel}, p_{t}^{act}, p_{t}^{un}] \in \mathbb{R}^5$, which denotes the probability that each word belongs to one of the $5$ categories. 
We calculate the global linguistic feature of actions $q_a \in \mathbb{R}^{C_l}$ by weighted combination of the all the words in the expression: 
\begin{equation} 
   \label{eq:act_feat}
   q_a = \sum\limits_{t=1}^{T}{p_t^{act}l_t}.
\end{equation}

Then, the video clip features $X_F$ are first fed into our entity perception stage to perceive all the possible entities in each frame independently and produce multimodal feature $M_V \in \mathbb{R}^{K \times H \times W \times C_m}$. 
Since the static relationships between entities in each frame are almost identical, we only select the annotated center frame of the video clip to conduct relation-aware reasoning stage to reduce the computational budgets. The reasoned feature of the center frame is denoted as $\bar{M}_V^{ctr} \in \mathbb{R}^{H \times W \times C_m}$.

As shown in Fig.~\ref{fig:cmpcv}, in order to highlight features of entities which are matched with the temporal action cues in the expression, we conduct cross-modal attention between $M_V$ and $q_a$ with necessary reshaping operations:
\begin{equation}
   \label{eq:tg_1}
   \hat{D}_V = (M_VW_8)(q_aW_9),
\end{equation}
\begin{equation}
   \label{eq:tg_2}
   D_V = softmax(\frac{\hat{D}_V}{\sqrt{C_m}}),
\end{equation}
\begin{equation}
   \label{eq:tg_3}
   P_V = D_VM_V,
\end{equation}
where $D_V \in \mathbb{R}^{K \times 1 \times HW}$, $M_V \in \mathbb{R}^{K \times HW \times C_m}$ after reshaping. We combine $K$ with the batch axis and perform a batch matrix-matrix product to obtain $P_V \in \mathbb{R}^{K \times 1 \times C_m}$.
Afterwards, we reshape $P_V$ to $\tilde{P}_V \in \mathbb{R}^{K \times C_m}$ and use $\tilde{P}_V$ as feature of vertexes to build a temporal graph $\mathcal{G_T}$ with $K$ frames as vertexes. 
Each vertex saves the global feature of entities matched with the action cues in each frame.
Then, we conduct matrix product between different transformations of $\tilde{P}_V \in \mathbb{R}^{K \times C_m}$ to obtain feature relevances among vertexes as the adjacency matrix $A_V \in \mathbb{R}^{K \times K}$ of $\mathcal{G_T}$: 
\begin{equation}
    \label{eq:tg_4}
    \hat{A}_V = (\tilde{P}_V W_{12})(\tilde{P}_V W_{13}),
\end{equation}
\begin{equation}
    \label{eq:tg_5}
    A_V = softmax(\frac{\hat{A}_V}{\sqrt{C_m}}).
\end{equation}
We further conduct graph convolution~\cite{kipf2016semi} among temporal vertexes as follows: 
\begin{equation}
    \label{eq:tg_6}
    \bar{P}_V = (A_V+I)\tilde{P}_VW_{14},
\end{equation}
The reasoned context $\bar{P}_V \in \mathbb{R}^{K \times C_m}$ is then projected to each spatial location of $\bar{M}_V^{ctr} \in \mathbb{R}^{H \times W \times C_m}$. 
We first conduct cross-modal attention between $\bar{P}_V$ and $\bar{M}_V^{ctr}$: 
\begin{equation}
    \label{eq:tg_7}
    \hat{E}_V = (\bar{P}_V W_{14})(\bar{M}_V^{ctr} W_{15}),
\end{equation}
\begin{equation}
    \label{eq:tg_8}
    E_V = softmax(\frac{\hat{E}_V}{\sqrt{C_m}}),
\end{equation}
where $E_V \in \mathbb{R}^{K \times HW}$ measures the feature relevances between $\bar{P}_V$ and $\bar{M}_V^{ctr}$. 
Then, we perform another matrix product between $E_V$ and $\bar{P}_V$ to obtain the image-format temporal context $\hat{P}_V \in \mathbb{R}^{H \times W \times C_m}$ with necessary reshaping operations: 
\begin{equation}
    \label{eq:tg_9}
    \hat{P}_V = {E_V}^T \bar{P}_V.
\end{equation}
Finally, we concatenate and fuse $M_V^{ctr}$, $\bar{M}_V^{ctr}$, $\hat{P}_V$ and repeated $s$ to produce feature $Y_V \in \mathbb{R}^{H \times W \times C_m}$ for the referent.

\begin{table*}[t]
   \centering
   \begin{tabular}{|m{2.2cm}<{\centering}|c|c|c|c|c|c|c|c|}
      \hline
      \multirow{2}*{Method} & \multicolumn{3}{c|}{UNC} & \multicolumn{3}{c|}{UNC+} & G-Ref & ReferIt \\
      \cline{2-9}
      & val & testA & testB & val & testA & testB & val & test \\
      \hline
      LSTM-CNN~\cite{hu2016segmentation} & - & - & - & - & - & - & 28.14 & 48.03 \\
      RMI~\cite{liu2017recurrent} & 45.18 & 45.69 & 45.57 & 29.86 & 30.48 & 29.50 & 34.52 & 58.73 \\
      DMN~\cite{margffoy2018dynamic} & 49.78 & 54.83 & 45.13 & 38.88 & 44.22 & 32.29 & 36.76 & 52.81 \\
      KWA~\cite{shi2018key} & - & - & - & - & - & - & 36.92 & 59.09 \\
      ASGN~\cite{qiu2019referring} & 50.46 & 51.20 & 49.27 & 38.41 & 39.79 & 35.97 & 41.36 & 60.31 \\
      RRN~\cite{li2018referring} & 55.33 & 57.26 & 53.95 & 39.75 & 42.15 & 36.11 & 36.45 & 63.63 \\
      MAttNet~\cite{yu2018mattnet} & 56.51 & 62.37 & 51.70 & 46.67 & 52.39 & 40.08 & n/a & - \\
      CMSA~\cite{ye2019cross} & 58.32 & 60.61 & 55.09 & 43.76 & 47.60 & 37.89 & 39.98 & 63.80 \\
      CAC~\cite{chen2019referring} & 58.90 & 61.77 & 53.81 & - & - & - & 44.32 & - \\
      STEP~\cite{chen2019see} & 60.04 & 63.46 & 57.97 & 48.19 & 52.33 & 40.41 & 46.40 & 64.13 \\
      \hline
      Ours-CVPR'20 & 61.36 & 64.53 & 59.64 & 49.56 & 53.44 & 43.23 & 49.05 & 65.53 \\
      Ours & \textbf{62.47} & \textbf{65.08} & \textbf{60.82} & \textbf{50.25} & \textbf{54.04} & \textbf{43.47} & \textbf{49.89} & \textbf{65.58} \\
      \hline
   \end{tabular}
   \caption{Comparison with state-of-the-art methods on four datasets for referring image segmentation. Overall IoU is adopted as the metric. ``n/a'' denotes MAttNet does not use the same split as other methods. $3$ rounds of feature exchange are adopted in TGFE.}
   \label{tab:sota_ris}
\end{table*}

\subsection{Text-Guided Feature Exchange}
As previous works~\cite{li2018referring}\cite{ye2019cross}\cite{chen2019see} demonstrate the importance of multi-level feature fusion in referring segmentation, 
we further introduce a Text-Guided Feature Exchange (TGFE) module which exploits visual and language context to communicate information among multi-level features. 
Our TGFE module takes visual features $Y^3, Y^4, Y^5$ (subscripts are omitted) and necessary sentence feature $s$ as inputs. 
After $n$ rounds of feature exchange, $Y_{n}^3, Y_{n}^4, Y_{n}^5$ are produced as outputs. 
For level $i$ at round $k$, TGFE module aims to exchange feature information from $Y_{k-1}^j, j \in \{3, 4, 5\} \backslash \{i\}$ to $Y_{k-1}^i$ and output aggregated feature $Y_{k}^i$. 
Concretely, we first conduct cross-modal attention between sentence feature $s \in \mathbb{R}^{C_l}$ and $Y_{k-1}^i \in \mathbb{R}^{H \times W \times C_m}$ with necessary reshaping and dimension transforming operations: 
\begin{equation}
   \label{eq:pool_weight}
   \Lambda_{k-1}^i = (s W_{10}) (Y_{k-1}^i W_{11})^T,
\end{equation}
where $\Lambda_{k-1}^i \in \mathbb{R}^{1 \times HW}$ measures the feature relevance between the whole sentence and each spatial location on $Y_{k-1}^i$. 
Then, we utilize $\Lambda_{k-1}^i$ as a weighted global pooling matrix to aggregate global context vector $g_{k-1}^i \in \mathbb{R}^{C_m}$ from $Y_{k-1}^i \in \mathbb{R}^{H \times W \times C_m}$: 
\begin{equation}
   \label{eq:global_vec}
   g_{k-1}^i = \Lambda_{k-1}^i Y_{k-1}^i.
\end{equation}
Then we fuse $s$ and $g_{k-1}^i$ with a fully connected layer to obtain a context vector $c_{k-1}^i \in \mathbb{R}^{C_m}$ which contains multimodal context of $Y_{k-1}^i$ with more textual information. 
Afterwards, we exploit $c_{k-1}^i$ to select features relevant to $Y_{k-1}^i$ from the other two levels of features $Y_{k-1}^j, j \in \{3, 4, 5\} \backslash \{i\}$ by channel attention. 
The original feature $Y_{k-1}^i$ of level $i$ at round $k - 1$ will be added with selected features from the other two levels at round $k - 1$ to obtain $Y_{k}^i$ of level $i$ at round $k$ as follows: 
\begin{equation}
   \label{eq:extract}
   Y_{k}^i = 
      \begin{cases}
         Y_{k-1}^i + \sum\limits_{j \in \{3, 4, 5\} \backslash \{i\}}{\sigma(c_{k-1}^i) \odot Y_{k-1}^j}, k \geq 1 \\
         \\
         Y^i, k = 0
      \end{cases}
\end{equation}
where $\sigma(\cdot)$ denotes sigmoid function. 
At each round, feature of each level $i \in \{3, 4, 5\}$ will select its relevant features from the other two levels under the guidance of textual information. 
After $n$ rounds of exchange, the output features $Y_{n}^3$, $Y_{n}^4$ and $Y_{n}^5$ are further fused with ConvLSTM~\cite{xingjian2015convolutional} to produce the final mask prediction. 

\section{Experiments}
\subsection{Experimental Setup}
\label{sec:setup}
\subsubsection{Datasets}
We conduct extensive experiments on four referring image segmentation benchmarks including UNC~\cite{yu2016modeling}, UNC+~\cite{yu2016modeling}, G-Ref~\cite{mao2016generation} and ReferIt~\cite{kazemzadeh2014referitgame}, and also on three referring video segmentation benchmarks including A2D Sentences~\cite{gavrilyuk2018actor}, J-HMDB Sentences~\cite{gavrilyuk2018actor} and Refer-Youtube-VOS~\cite{seo2020urvos}.

UNC, UNC+ and G-Ref are all collected on MS-COCO~\cite{lin2014microsoft}. They contain $19,994$, $19,992$ and $26,711$ images with $142,209$, $141,564$ and $104,560$ referring expressions for over $50,000$ objects, respectively. 
Expressions in UNC+ contain no location words while those in G-Ref have much longer length than others. 
ReferIt is collected on IAPR TC-12~\cite{escalante2010segmented} and contains 
$19,894$ images with $130,525$ expressions for $96,654$ objects (including stuff).
A2D Sentences is extended from the Actor-Action Dataset~\cite{xu2015can} by providing textual descriptions for each video. 
It contains $3,782$ videos annotated with $8$ action classes performed by $7$ actor classes.
J-HMDB sentences is extended from the J-HMDB dataset~\cite{jhuang2013towards} which contains $21$ different actions, $928$ videos and corresponding $928$ sentences. All the actors in JHMDB dataset are humans and one natural language query is annotated to describe the action performed by each actor.
Refer-Youtube-VOS is a large-scale referring video segmentation dataset extended from Youtube-VOS dataset~\cite{xu2018youtube} which contains $3975$ videos, $7451$ objects and $27899$ expressions with both first-frame expression and full-video expression annotated.

\subsubsection{Evaluation Metrics}
We adopt Prec@X and overall Intersection-over-Union (overall IoU) as metrics to evaluate our image model. 
Prec@X measures the percentage of test samples whose IoU with ground-truth masks are higher than the threshold $X \in \{0.5, 0.6, 0.7, 0.8, 0.9\}$. 
Overall IoU accumulates the total intersection regions over total union regions of all the test samples. 
For video model, we additionally use mean Average Precision (mAP) and mean IoU as metrics in addition to Prec@X and overall IoU.

\subsubsection{Implementation Details}
We adopt DeepLab-ResNet101~\cite{chen2017deeplab} which is pretrained on the PASCAL-VOC dataset~\cite{everingham2010pascal} as the CNN backbone to extract visual features for the input image and video.
The output of Res$3$, Res$4$ and Res$5$ are used for multi-level feature fusion. Input images and video frames are resized to $320 \times 320$.
The input video clip contains $5$ frames for CMPC-v model. 
Besides, for video segmentation, we further adopt I3D~\cite{carreira2017quo} pretrained on Kinetics dataset~\cite{kay2017kinetics} as backbone of CMPC-v model to compare fairly with previous methods~\cite{wang2019asymmetric,ning2020polar}. 
Due to the temporal downsamping operation in I3D, the input video clip of our model contains $8$ frames following PRPE to retain temporal information throughout the network. 
The output of last three stages of I3D are used for multi-level feature fusion.
Channel dimensions of features are set as $C_v = C_l = C_m = C_h = 1000$ and the cell size of ConvLSTM~\cite{xingjian2015convolutional} is set to $500$.
When comparing with other methods, the hyper-parameter $r$ of bilinear fusion is set to $5$ and the number of feature exchange rounds $n$ is set to $3$. 
GloVe word embeddings~\cite{pennington2014glove} 
pretrained on Common Crawl 840B tokens are adopted following~\cite{chen2019see}. 
Number of graph convolution layers is set as $2$ on G-Ref dataset and $1$ on others.
We train the network using Adam optimizer~\cite{kingma2014adam} with the initial learning rate of $2.5e^{-4}$ and weight decay of $5e^{-4}$. 
Parameters of CNN backbone are fixed during training. 
Binary cross-entropy loss averaged over all pixels is used for training. 
DenseCRF~\cite{krahenbuhl2011efficient} is adopted to refine the segmentation masks for fair comparison with prior works.

\begin{table*}[!htbp]
   \centering
   \begin{tabular}{|m{4.2cm}<{\centering}|c|c|c|c|c|c|c|c|}
         \hline
         \multirow{2}*{Method} & \multicolumn{5}{c|}{Precision} & mAP & \multicolumn{2}{c|}{IoU} \\
         \cline{2-9}
         & Prec@0.5 & Prec@0.6 & Prec@0.7 & Prec@0.8 & Prec@0.9 & 0.5:0.95 & Overall & Mean \\
         \hline
         Hu \textit{et al.}~\cite{hu2016segmentation} & 34.8 & 23.6 & 13.3 & 3.3 & 0.1 & 13.2 & 47.4 & 35.0 \\
         Li \textit{et al.}~\cite{li2017tracking} & 38.7 & 29.0 & 17.5 & 6.6 & 0.1 & 16.3 & 51.5 & 35.4 \\
         Gavrilyuk \textit{et al.}~\cite{gavrilyuk2018actor} (RGB) & 47.5 & 34.7 & 21.1 & 8.0 & 0.2 & 19.8 & 53.6 & 42.1 \\
         Gavrilyuk \textit{et al.}~\cite{gavrilyuk2018actor} (RGB+Flow) & 50.0 & 37.6 & 23.1 & 9.4 & 0.4 & 21.5 & 55.1 & 42.6 \\
         ACGANet~\cite{wang2019asymmetric} (RGB) & 55.7 & 45.9 & 31.9 & 16.0 & 2.0 & 27.4 & 60.1 & 49.0 \\
         PRPE~\cite{ning2020polar} & 63.4 & 57.9 & 48.3 & 32.2 & 8.3 & 38.8 & \textbf{66.1} & 52.9 \\
         \hline
         Ours-R2D & 59.0 & 52.7 & 43.4 & 28.4 & 6.8 & 35.1 & 64.9 & 51.5 \\
         Ours-I3D & \textbf{65.5} & \textbf{59.2} & \textbf{50.6} & \textbf{34.2} & \textbf{9.8} & \textbf{40.4} & 65.3 & \textbf{57.3} \\
         \hline
   \end{tabular}
   \caption{Comparison with state-of-the-art methods on A2D Sentences for referring video segmentation. Our method significantly outperforms state-of-the-arts using only RGB input.}
   \label{tab:sota_rvs_a2d}
\end{table*}

\begin{table*}[!htbp]
   \centering
   \begin{tabular}{|m{4.2cm}<{\centering}|c|c|c|c|c|c|c|c|}
         \hline
         \multirow{2}*{Method} & \multicolumn{5}{c|}{Precision} & mAP & \multicolumn{2}{c|}{IoU} \\
         \cline{2-9}
         & Prec@0.5 & Prec@0.6 & Prec@0.7 & Prec@0.8 & Prec@0.9 & 0.5:0.95 & Overall & Mean \\
         \hline
         Gavrilyuk \textit{et al.}~\cite{gavrilyuk2018actor} (RGB+Flow) & 69.9 & 46.0 & 17.3 & 1.4 & 0.0 & 23.3 & 54.1 & 54.2 \\
         ACGANet~\cite{wang2019asymmetric} (RGB) & 75.6 & 56.4 & 28.7 & 3.4 & 0.0 & 28.9 & 57.6 & 58.4 \\
         PRPE~\cite{ning2020polar} & 69.0 & 57.2 & 31.9 & 6.0 & \textbf{0.1} & 29.4 & - & - \\
         \hline
         Ours (RGB) & \textbf{81.3} & \textbf{65.7} & \textbf{37.1} & \textbf{7.0} & 0.0 & \textbf{34.2} & \textbf{61.6} & \textbf{61.7} \\
         \hline
   \end{tabular}
   \caption{Comparison with state-of-the-art methods on JHMDB Sentences for referring video segmentation. Our method significantly outperforms previous ones using only RGB input.}
   \label{tab:sota_J-HMDB}
\end{table*}

\begin{table*}[h]
   \centering
   \begin{tabular}{|m{5cm}<{\centering}|c|c|c|c|c|c|c|}
         \hline
         \multirow{2}*{Method} & \multirow{2}*{$\mathcal{J} $} & \multirow{2}*{$\mathcal{F} $} & \multicolumn{5}{c|}{Precision} \\
         \cline{4-8}
         &  &  & Prec@0.5 & Prec@0.6 & Prec@0.7 & Prec@0.8 & Prec@0.9 \\
         \hline
         URVOS w/o memory attention~\cite{seo2020urvos} & 39.38 & 41.78 & 46.26 & 40.98 & 34.81 & 25.42 & 10.86 \\
         URVOS~\cite{seo2020urvos} & 45.27 & 49.19 & \textbf{52.19} & 46.77 & 40.16 & 27.67 & 14.11 \\
         \hline
         Ours & \textbf{45.64} & \textbf{49.32} & 51.65 & \textbf{47.16} & \textbf{40.96} & \textbf{31.21} & \textbf{16.52}  \\
         \hline
   \end{tabular}
   \caption{Comparison with state-of-the-art methods on Refer-Youtube-VOS dataset. Our method outpreforms previous ones without memory attention across frames.}
   \label{tab:sota_Refer-Youtube-VOS}
\end{table*}

\begin{table*}[t]
  \centering
  \begin{tabular}{|l|cccccc|ccccc|c|}
      \hline
      & EP & RAR & TGFE & CLSTM & GloVe & CMF & Prec@0.5 & Prec@0.6 & Prec@0.7 & Prec@0.8  & Prec@0.9  & Overall IoU \\
      \hline
      1 & & & & & & & 48.01 & 37.98 & 27.92 & 16.30 & 3.72 & 47.36 \\
      2 & $\surd$ & & &  & & & 49.76 & 40.35 & 30.15 & 17.84 & 4.16 & 49.06 \\
      3 & & $\surd$ & & &  & & 59.32 & 51.16 & 40.59 & 26.50 & 6.66 & 53.40 \\
      4 & $\surd$ & $\surd$ & & & & & 62.86 & 54.54 & 44.10 & 28.65 & 7.24 & 55.38 \\
      5 & $\surd$ & $\surd$ & & & $\surd$ & & 62.87 & 54.91 & 44.16 & 28.43 & 7.23 & 56.00 \\
      6 & $\surd$ & $\surd$ & & & $\surd$ & $\surd$ & \textbf{65.77} & \textbf{57.80} & \textbf{47.13} & \textbf{31.00} & \textbf{7.75} & \textbf{57.12} \\
      \hline
      7 & & & & $\surd$ & &  & 63.12 & 54.56 & 44.20 & 28.75 & 8.51 & 56.38 \\
      8 & & & $\surd$ & $\surd$ & & & 67.63 & 59.80 & 49.72 & 34.45 & 10.62 & 58.81 \\
      9 & $\surd$ & & $\surd$ & $\surd$ & & & 68.39 & 60.92 & 50.70 & 35.24 & 11.13 & 59.05 \\
      10 & & $\surd$ & $\surd$ & $\surd$ & & & 69.37 & 62.28 & 52.66 & 36.89 & 11.27 & 59.62 \\
      11 & $\surd$ & $\surd$ & $\surd$ & $\surd$ & & & 71.04 & 64.02 & 54.25 & 38.45 & 11.99 & 60.72 \\
      12 & $\surd$ & $\surd$ & $\surd$ & $\surd$ & $\surd$ & & 71.27 & 64.44 & 55.03 & 39.28 & \textbf{12.89} & 61.19 \\
     13 & $\surd$ & $\surd$ & & $\surd$ & $\surd$  & $\surd$ & 70.79 & 63.08 & 53.00 & 36.89 & 11.27 & 60.43 \\
      14 & $\surd$ & $\surd$ & $\surd$ & & $\surd$ & $\surd$ & 72.01 & 64.90 & 55.37 & 39.03 & 12.12 & 61.16 \\
      15 & $\surd$ & $\surd$ & $\surd$ & $\surd$ & $\surd$ & $\surd$ & \textbf{72.08} & \textbf{65.30} & \textbf{55.65} & \textbf{39.74} & 12.80 & \textbf{61.80} \\
      \hline
  \end{tabular}
  \caption{Ablation studies of CMPC-I and TGFE modules on UNC val set.
  EP and RAR indicate entity perception and relation-aware reasoning stages. CMF denotes concatenating multimodal feature from EP instead of pure visual feature for the output of CMPC-I. Here TGFE adopts one round of feature exchange.}
  \label{tab:cmpc_i}
\end{table*}

\subsection{Comparison with State-of-the-arts}
\subsubsection{Referring Image Segmentation}
To demonstrate the superiority of our method for referring image segmentation, we evaluate it on four benchmark datasets. 
As illustrated in Table~\ref{tab:sota_ris}, our method outperforms previous state-of-the-arts on all the datasets with large margins. 
Comparing with STEP~\cite{chen2019see} which densely fuses $5$ levels of features for $25$ times, our method utilizes fewer levels of features and fusion times while consistently obtaining $1.45\%$-$3.49\%$ performance gains on all the four datasets, demonstrating the effectiveness of our modules. 
Particularly, our method yields $3.49\%$ IoU improvement over STEP on G-Ref val set, indicating our method can better handle long sentences with progressive comprehension. 
Besides, ReferIt is a challenging dataset and previous methods only obtain marginal improvements on it. 
For instance, STEP and CMSA~\cite{ye2019cross} achieve only $0.33\%$ and $0.17\%$ improvements on ReferIt test set respectively, while our method enlarges the performance gain to $1.45\%$, which shows that our model can well segment both objects and stuff.
In addition, our method also outperforms MAttNet~\cite{yu2018mattnet} by a large margin in overall IoU. 
MAttNet depends on Mask R-CNN~\cite{he2017mask} pretrained on much more COCO~\cite{lin2014microsoft} images ($110K$) than ours pretrained on PASCAL-VOC~\cite{everingham2010pascal} images ($10K$) to generate RoI proposals. 
Therefore, it may not be completely fair to directly compare performances of MAttNet with ours.

\subsubsection{Referring Video Segmentation}
Comparisons with state-of-the-art methods on A2D Sentences dataset are summarized in Table~\ref{tab:sota_rvs_a2d}.
Since prior works~\cite{gavrilyuk2018actor, wang2019asymmetric, ning2020polar} adopt I3D~\cite{carreira2017quo} as visual backbone to encode video features, we also build a 3D-version of our CMPC-V network for fair comparison.
We present results of both 2D backbone and 3D backbone in Table~\ref{tab:sota_rvs_a2d}, which are denoted as `Ours-R2D' and `Ours-I3D' respectively. 
Our I3D-based model achieves notable improvements comparing with the R2D-based model, indicating that 3D backbone extracts more temporal information. 
Our method also outperforms previous state-of-the-arts, PRPE~\cite{ning2020polar},
on most evaluation metrics except Overall IoU, where our model achieves comparable result with PRPE.

To further demonstrate the generalization ability of our video model, we conduct experiments on the J-HMDB Sentences dataset~\cite{gavrilyuk2018actor} and the Refer-Youtube-VOS dataset~\cite{seo2020urvos}. 
We follow prior works~\cite{gavrilyuk2018actor, wang2019asymmetric, ning2020polar} to use the best model pretrained on A2D Sentences dataset to directly evaluate all the test samples of J-HMDB Sentences dataset without finetuning. 
As shown in Table \ref{tab:sota_J-HMDB}, our video model achieves notable performance gain over previous methods for most evaluation metrics ($4.0\%$ Overall IoU, $5.7\%$ Prec@$0.5$, etc.), indicating that our method obtains stronger generalization ability. 
Please note that all the methods including ours produce $0.0\%$ or $0.1\%$ on Prec@$0.9$, which is probably because without training on J-HMDB Sentences, models cannot generate particularly fine masks on unseen samples.

For Refer-Youtube-VOS dataset, we train our video model for $200,000$ iterations with $5e^{-4}$ as initial learning rate (divided by $10$ at $160,000$th iteration). As shown in Table~\ref{tab:sota_Refer-Youtube-VOS}, Our video model outperforms URVOS~\cite{seo2020urvos} on most metrics except Prec@$0.5$ without further refining segmentation masks using memory attention between frames. 
The comparison shows that our model is able to recognize the referred objects without too much interactions among frames.

\subsection{Ablation Studies}
We perform ablation studies on UNC val set, G-Ref val set and A2D Sentence test set to testify the effectiveness of each proposed module for referring image and video segmentation.

\begin{table*}[t]
   \centering
   \begin{tabular}{|l|c|c|c|c|c|c|c|c|c|c|c|c|}
      \hline
      & \multirow{2}*{TGFE*} & \multirow{2}*{EP} & \multirow{2}*{RAR} & \multirow{2}*{AAR} & \multicolumn{5}{c|}{Precision} & mAP & \multicolumn{2}{c|}{IoU} \\
      \cline{6-13}
      & & & & & Prec@0.5 & Prec@0.6 & Prec@0.7 & Prec@0.8 & Prec@0.9 & 0.5:0.95 & Overall & Mean \\
      \hline
      1 & & & & & 52.8 & 47.2 & 38.7 & 25.4 & 5.9 & 31.3 & 62.3 & 46.6 \\
      2 & $\surd$ & & & & 56.7 & 49.7 & 40.3 & 25.6 & 5.8 & 32.7 & 63.2 & 49.3 \\
      3 & $\surd$ & $\surd$ & & & 56.5 & 50.3 & 40.7 & 26.9 & 6.2 & 33.1 & 63.9 & 49.8 \\
      4 & $\surd$ & $\surd$ & $\surd$ & & 57.6 & 50.2 & 40.2 & 26.1 & 5.8 & 33.1 & 63.2 & 49.7 \\
      5 & $\surd$ & $\surd$ & $\surd$ & $\surd$ & \textbf{59.0} & \textbf{52.7} & \textbf{43.4} & \textbf{28.4} & \textbf{6.8} & \textbf{35.1} & \textbf{64.9} & \textbf{51.5} \\
      \hline
   \end{tabular}
   \caption{Ablation studies of our CMPC-V modules on A2D Sentence test set. 
   EP, RAR and AAR indicate entity perception, relation-aware reasoning and action-aware reasoning stages in our CMPC-V module. GloVe and CMF are adopted in the baseline by default for clarity of presentation. TGFE* here means TGFE together with ConvLSTM.}
   \label{tab:cmpc_v}
\end{table*}

\subsubsection{Components of CMPC-I Module}
We first explore the effectiveness of each component of our proposed CMPC-I module. 
Experimental results are summarized in Table~\ref{tab:cmpc_i}.
EP and RAR denotes the entity perception and relation-aware reasoning stages in CMPC-I module respectively. 
GloVe means using GloVe word embeddings~\cite{pennington2014glove} to initialize the parameters of embedding layer. 
CMF means concatenating multimodal feature from EP instead of pure visual feature to produce the output of CMPC-I, as mentioned in the last paragraph of Section~\ref{sec:rar}. 
Results in rows $1$ to $6$ are all based on single-level visual feature, i.e. Res$5$. 
Our baseline (row $1$) simply concatenates the visual feature from DeepLab-$101$ and linguistic feature from LSTM and makes predictions on the fusion of them.

As shown in row $2$ of Table~\ref{tab:cmpc_i}, including EP brings $1.70\%$ IoU improvement over the baseline, indicating the perception of candidate entities can help model to eliminate noisy backgrounds. 
In row $3$, RAR alone brings $6.04\%$ IoU improvement over baseline, demonstrating that the referent can be effectively highlighted by leveraging relational words as routers to reason among spatial regions, thus boosting the performance notably. 
Combining EP with RAR, our CMPC-I module can achieve $55.38\%$ IoU with single level visual feature, enlarging the performance margin to $8.02\%$ IoU. 
This indicates that our model can accurately identify the referent by progressively comprehending the input image and expression. 
Integrated with GloVe word embeddings, the IoU gain achieves $8.64\%$ with the aid of large-scale corpus. 
As shown in row $13$ ad $14$, our proposed TGFE has significant influence on the performance while ConvLSTM only yields marginal improvements, demonstrating the effectiveness of TGFE.
Particularly, CMF further boosts the performance gain to $9.76\%$, which shows concatenating multimodal feature can provide richer context than pure visual feature. 

We further conduct ablation studies based on multi-level visual features as shown in rows $7$ to $13$ of Table~\ref{tab:cmpc_i}. 
Row $7$ is the multi-level version of row $1$ using ConvLSTM to fuse multi-level features. 
The TGFE module in rows $7$ to $11$ adopts single round of feature exchange. 
As shown in Table~\ref{tab:cmpc_i}, our model yields consistent performance gains with the single-level version, which demonstrates the effectiveness of our CMPC-I module under multi-level situation.

\subsubsection{Components of CMPC-V Module}
We build the CMPC-V module based on CMPC-I module by introducing an additional action-aware reasoning (AAR) stage to exploit temporal information for identifying the referent. 
Our video-version baseline contains GloVe and CMF by default and we evaluate TGFE, EP, RAR and AAR respectively for clarity. 
The experimental results are summarized in Table~\ref{tab:cmpc_v}. 
We can observe that TGFE can bring notable gains on most of metrics benefited from multi-level visual features. 
Combining EP and RAR stages can improve the performance slightly by utilizing only spatial information. 
It should be noticed that incorporating our proposed AAR stage is able to further obtain large performance gain over our strong baseline using TGFE, EP and RAR stages, demonstrating that temporal context information is critical to the referring video segmentation task.

We also tried to use action words as routers to obtain the adjacency matrix of AAR as in RAR, and the results are shown in Table \ref{tab:adj_mat_aar}. 
AR and DR represent ``Adaptive Relevance'' in RAR and ``Direct Relevance'' in original AAR respectively. 
Adaptive relevance yields inferior performance than direct relevance, indicating that direct relevance is more suitable to propagate information among different frames.

\begin{table}[!htbp]
   \centering
   \begin{tabular}{|m{2.2cm}<{\centering}|c|c|c|}
         \hline
         \multirow{2}*{Method} & mAP & \multicolumn{2}{c|}{IoU} \\
         \cline{2-4}
         & 0.5:0.95 & Overall & Mean \\
         \hline
         AR & 34.2 & 64.5 & 49.9 \\
         DR (Ours) & \textbf{35.1} & \textbf{64.9} & \textbf{51.5} \\
         \hline
   \end{tabular}
   \caption{Ablation studies of different ways to obtain adjacency matrix in AAR on A2D dataset. \textit{AR} means ``Adaptive Relevance'', which denotes the adjacency matrix is obtained in the similar way of RAR. \textit{DR} means ``Direct Relevance'', which denotes the adjacency matrix is obtained with direct feature relevance in our original implementation.}
   \label{tab:adj_mat_aar}
\end{table}

\subsubsection{TGFE module}
Table~\ref{tab:tgfe} presents the ablation results of TGFE module for referring image segmentation. 
$n$ is the number of feature exchange rounds. 
Our experiments are conducted upon CMPC-I module without CMF. 
Results show that only one round of feature exchange in TGFE could improve the IoU from $59.85\%$ to $60.72\%$. 
The IoU performance increases as the number of feature exchange rounds increases, which well proves the effectiveness of our TGFE module. 
We further directly incorporate TGFE module with the baseline model and results are shown in row $7$ and row $8$ of Table~\ref{tab:cmpc_i}.
TGFE with single round of feature exchange boosts the IoU from $56.38\%$ to $58.81\%$, indicating our TGFE module can effectively utilize rich contexts in multi-level features.

We  also tried to remove the additional necessary words extracting process in TGFE stage and the results are shown in Table~\ref{tab:nw}. 
Results shows that extracting necessary words features can yield slight improvements. It indicates the words extraction is not redundant.

\begin{table}[!htbp]
   \centering
   \begin{tabular}{|c|c|c|c|}
      \hline
      \multirow{2}*{CMPC-I only} & \multicolumn{3}{c|}{+TGFE} \\
      \cline{2-4}
      & $n=1$ & $n=2$ & $n=3$ \\
      \hline
      59.85 & 60.72 & 61.07 & \textbf{61.25} \\
      \hline
   \end{tabular}
   \caption{Overall IoUs of different numbers of feature exchange rounds in TGFE module on UNC val set. $n$ denotes the number of feature exchange rounds. \textit{Remove the engineering technique CMF from CMPC-I module.}}
   \label{tab:tgfe}
\end{table}

\begin{table}[!htbp]
   \centering
   \begin{tabular}{|c|ccc|c|}
     \hline
     Method & Prec@0.5 & Prec@0.7 & Prec@0.9  & Overall IoU \\
     \hline
     Full Model & \textbf{72.08} & \textbf{55.65} & \textbf{12.80} & \textbf{61.80} \\
     \hline
     w/o NW & 71.57 & 56.11 & 12.51 & 61.23 \\
     \hline
   \end{tabular}
   \caption{Ablation studies of necessary words extracting (NW) on UNC val set.}
   \label{tab:nw}
\end{table}

\subsubsection{Number of Graph Convolution Layer}
In Table~\ref{tab:gcn}, we explore the number of graph convolution layers in CMPC-I module based on single-level feature without CMF. 
$g$ denotes the number of graph convolution layers in CMPC-I.
Results on UNC val set show that more graph convolution layers lead to performance degradation.
However, on G-Ref val set, $2$ layers of graph convolution in CMPC-I obtains better performance than $1$ layer while $3$ layers decreasing the performance. 
Since G-Ref has much longer expressions (average length of $8.4$ words) than UNC (average length $< 4$ words), we suppose that stacking more graph convolution layers in CMPC-I can appropriately improve the reasoning ability for longer referring expressions. 
However, too many graph convolution layers may introduce noises and harm the performance.

\begin{table}[!htbp]
   \centering
   \begin{tabular}{|c|c|c|c|c|}
         \hline
         \multirow{2}*{Dataset}  & \multicolumn{4}{c|}{CMPC-I} \\
         \cline{2-5}
         & $g=0$ & $g=1$ & $g=2$ & $g=3$ \\
         \hline
         UNC val & 49.06 & \textbf{55.38} & 51.57 & 50.70 \\
         \hline
         G-Ref val & 36.50 & 38.19 & \textbf{40.12} & 38.96 \\
         \hline
   \end{tabular}
   \caption{Experiments of graph convolution on UNC val set and G-Ref val set in terms of \textit{overall IoU}. 
   $g$ denotes the number of graph convolution layers. 
   Experiments are all conducted on single level feature w/o CMF.}
   \label{tab:gcn}
\end{table}

\begin{table*}[!htbp]
   \centering
   \begin{tabular}{|m{2.2cm}<{\centering}|c|c|c|c|c|c|c|c|}
      \hline
      \multirow{2}*{Method} & \multicolumn{3}{c|}{UNC} & \multicolumn{3}{c|}{UNC+} & G-Ref & ReferIt \\
      \cline{2-9}
      & val & testA & testB & val & testA & testB & val & test \\
      \hline
      Random  & 57.39 & 59.65 & 55.86 & 46.94 & 50.95 & 40.66 & 47.28 & 61.49 \\
      Ours & \textbf{62.47} & \textbf{65.08} & \textbf{60.82} & \textbf{50.25} & \textbf{54.04} & \textbf{43.47} & \textbf{49.89} & \textbf{65.58} \\
      \hline
   \end{tabular}
   \caption{Comparison with random word type assignment on 4 datasets for referring image segmentation. Overall IoU is adopted as the metric.}
   \label{tab:random}
\end{table*}

\subsection{Visualization Analyses}
\label{sec:visualization}

\subsubsection{Accuracy of Word Classification}
\begin{figure}[htbp]
   \centering 
   \includegraphics[width=1.00\linewidth]{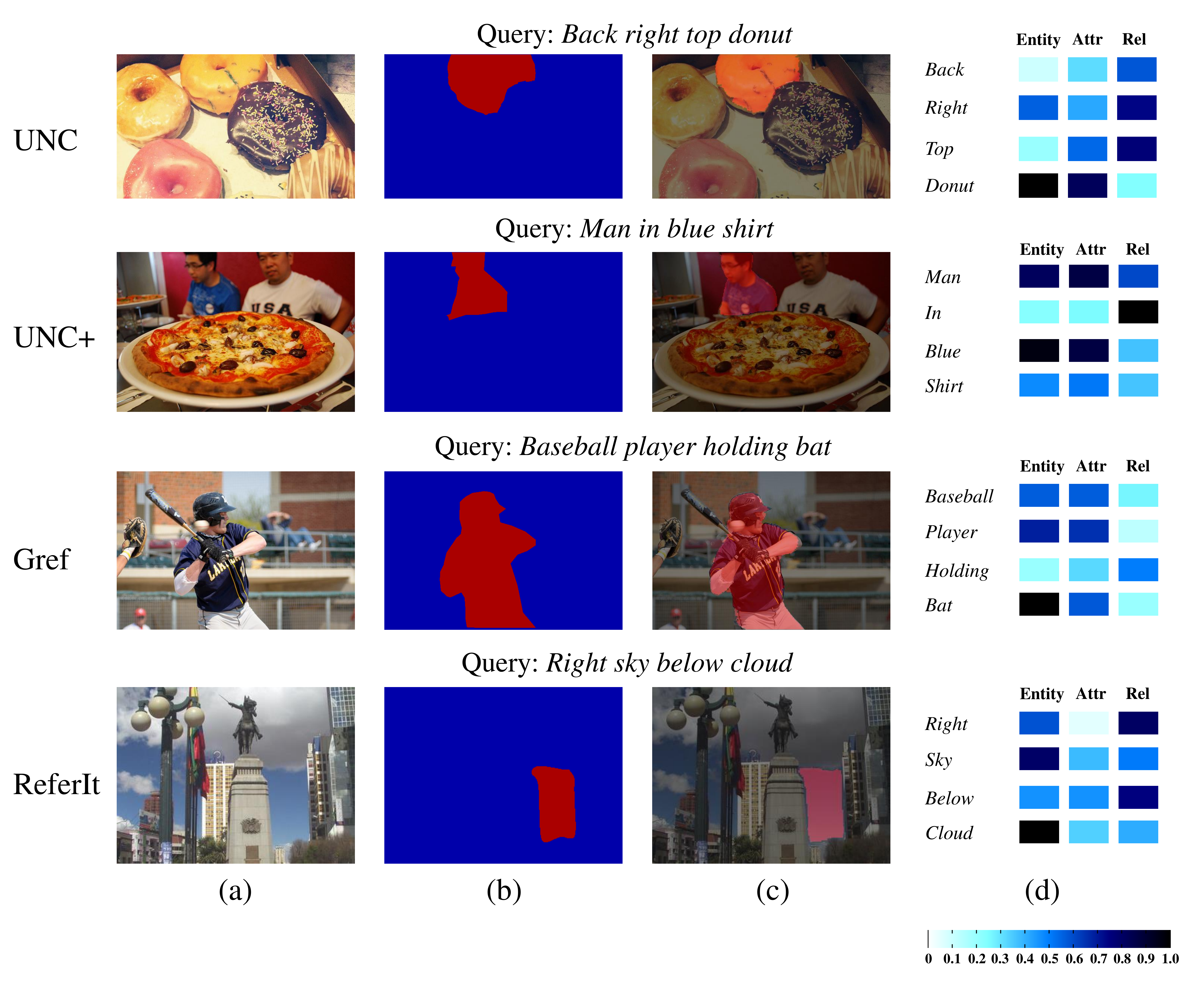}
   \caption{Visualization of words classification probabilities on four benchmark datasets. (a) Original image. (b) Ground-truth. (c) Prediction of our model. (d) Word classification probabilities of our model. Types include Entity, Attribute (Attr) and Relation (Rel). Darker color denotes larger probability.}
   \label{fig:word_attn}
 \end{figure}

As mentioned in Section~\ref{sec:ep}, we supervise the learning of word classification by the final segmentation loss due to the lack of annotations for word types.
The $4$ types of words, i.e., entity, attribute, relation and unnecessary, are implicitly defined according to the role each word plays in the whole expression and it is hard to quantitatively evaluate words classification accuracy. 
Thus we show the visualization of the words classification probabilities which are predicted by our model in Fig.~\ref{fig:word_attn} (d).
Among the three blue blocks, from left to right are the probabilities of the word being entity, attribute and relation types where darker color denotes larger probability. 
In the first example of expression “back right top donut”, words “back”, “right” and “top” have largest probabilities of being relation type, while word “donut” has largest probability of being entity type. 
This indicates our model can well recognize the type of each word in a soft manner and further utilize these words to highlight features of the referent by our CMPC module.

\subsubsection{Correlations of Word Classification and Segmentation}
To demonstrate our model can learn meaningful word classification results, 
we randomly set the classification probabilities for words in the expression during testing.
Experimental results are summarized in Table~\ref{tab:random}. 
Assigning random classification probabilities to each word leads to notable performance degradation, 
which shows the implicit learning of word classification is able to guide the progressive comprehension of referring expressions.

\begin{figure}[!htbp]
    \begin{center}
       \includegraphics[width=1.0\linewidth]{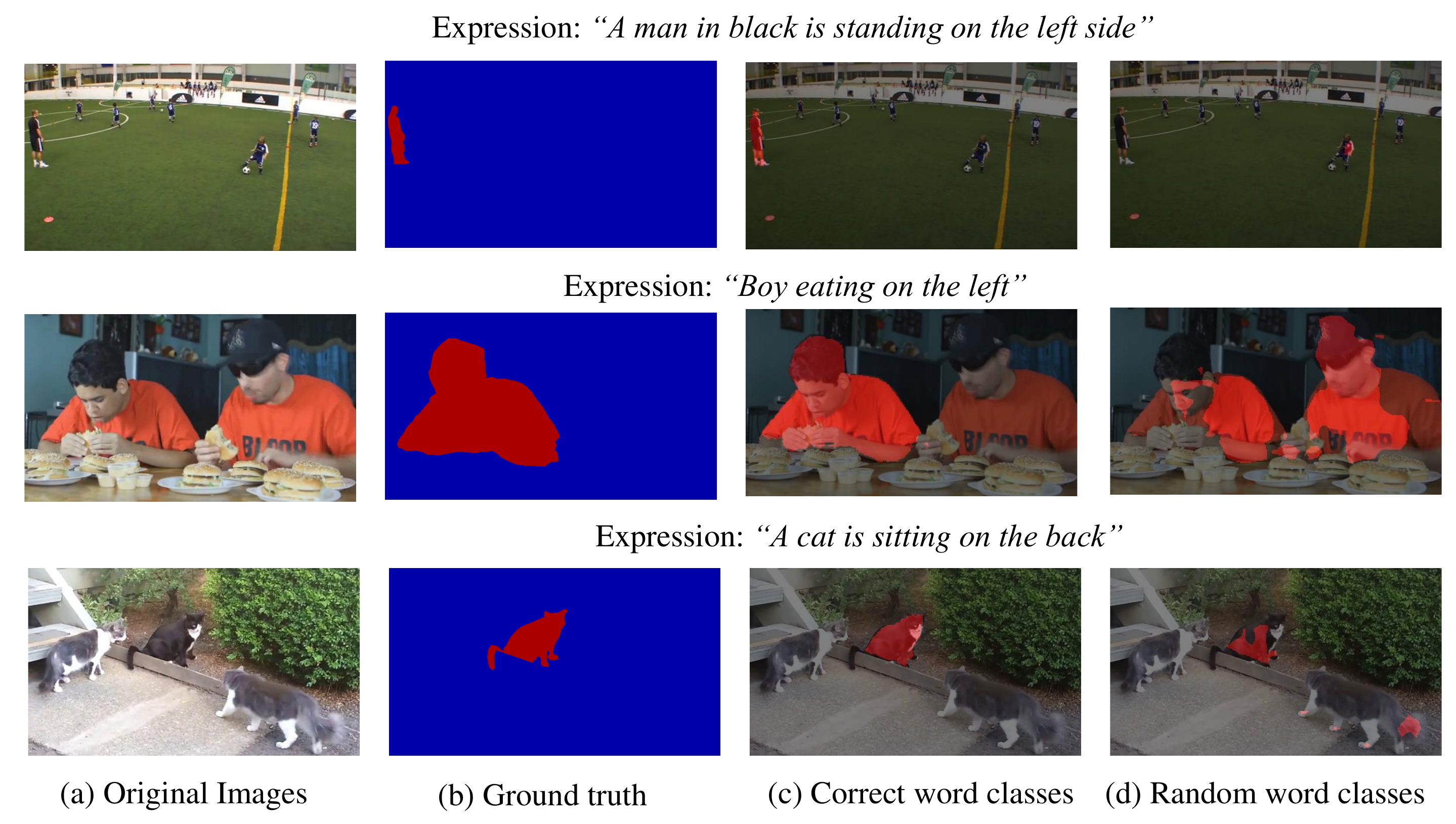}
    \end{center}
       \caption{Segmentation results of random word type assignment. (a) Original images. (b) Ground truth masks for referred objects. (c) Segmentation results of our model with correct word categories. (d) Segmentation results of our model with randomly assigned word categories.}
    \label{fig:wrong}
 \end{figure}

We also visualized the segmentation results with random word classification probabilities in the expressions during testing in Fig.~\ref{fig:wrong}. 
The segmentation results show that with randomly assigned word categories, the model cannot identify the correct object described in the expression. 
Taking the first row as an example, after modifying the word categories, the model mis-recognizes the man in the middle of the image as referred object, 
while model with original word categories could make correct prediction. 
These results show that our model can learn meaningful word classification results without direct supervision.

 \begin{figure}[!htbp]
    \begin{center}
       \includegraphics[width=0.85\linewidth]{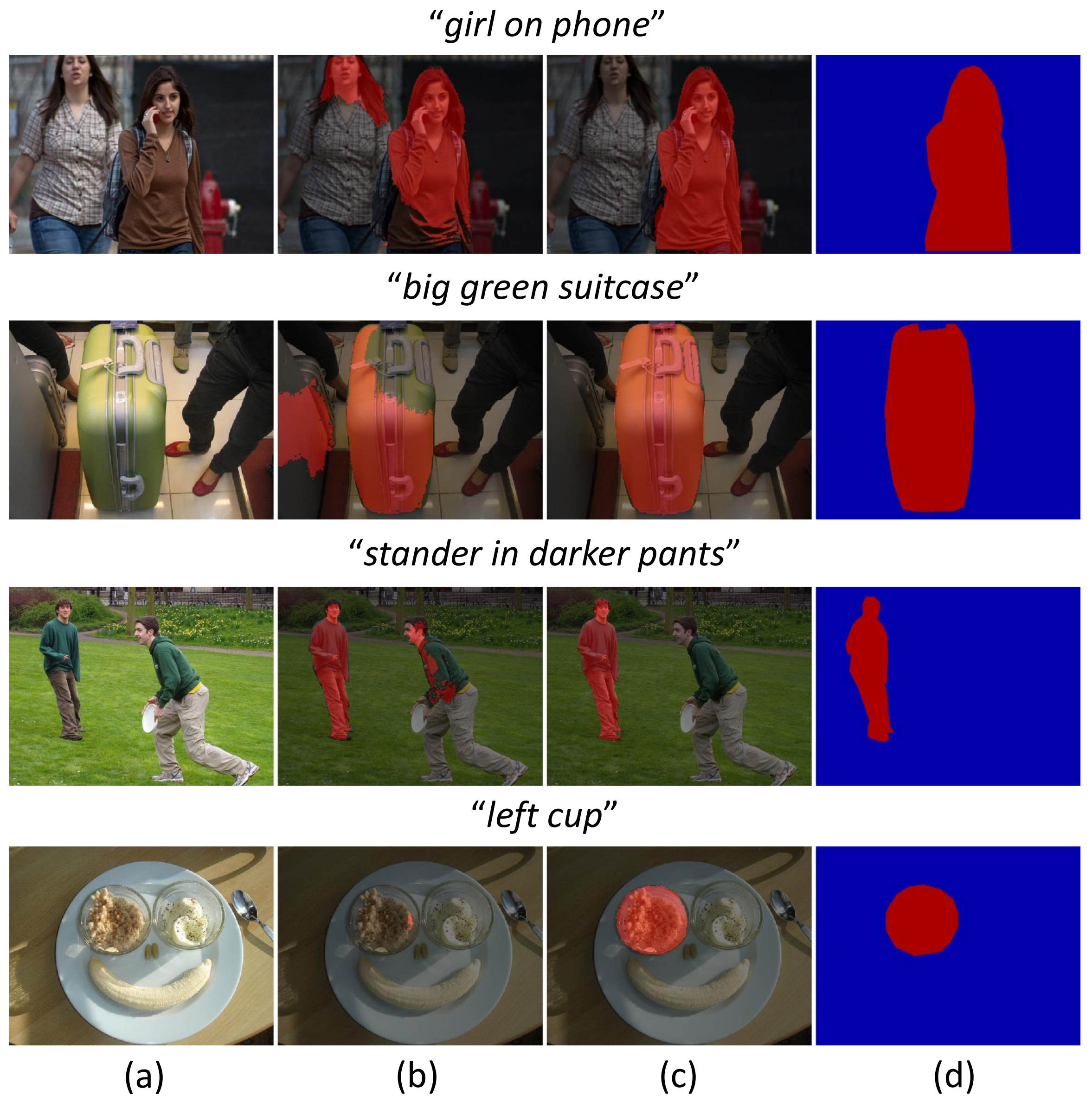}
    \end{center}
       \caption{Qualitative results of referring image segmentation. (a) Original image. (b) Results of the multi-level baseline model (row 7 in Table~\ref{tab:cmpc_i}). (c) Results of our model (row 12 in Table~\ref{tab:cmpc_i}). (d) Ground-truth.}
    \label{fig:quality-i}
 \end{figure}
 
 \begin{figure}[!htbp]
    \begin{center}
       \includegraphics[width=0.85\linewidth]{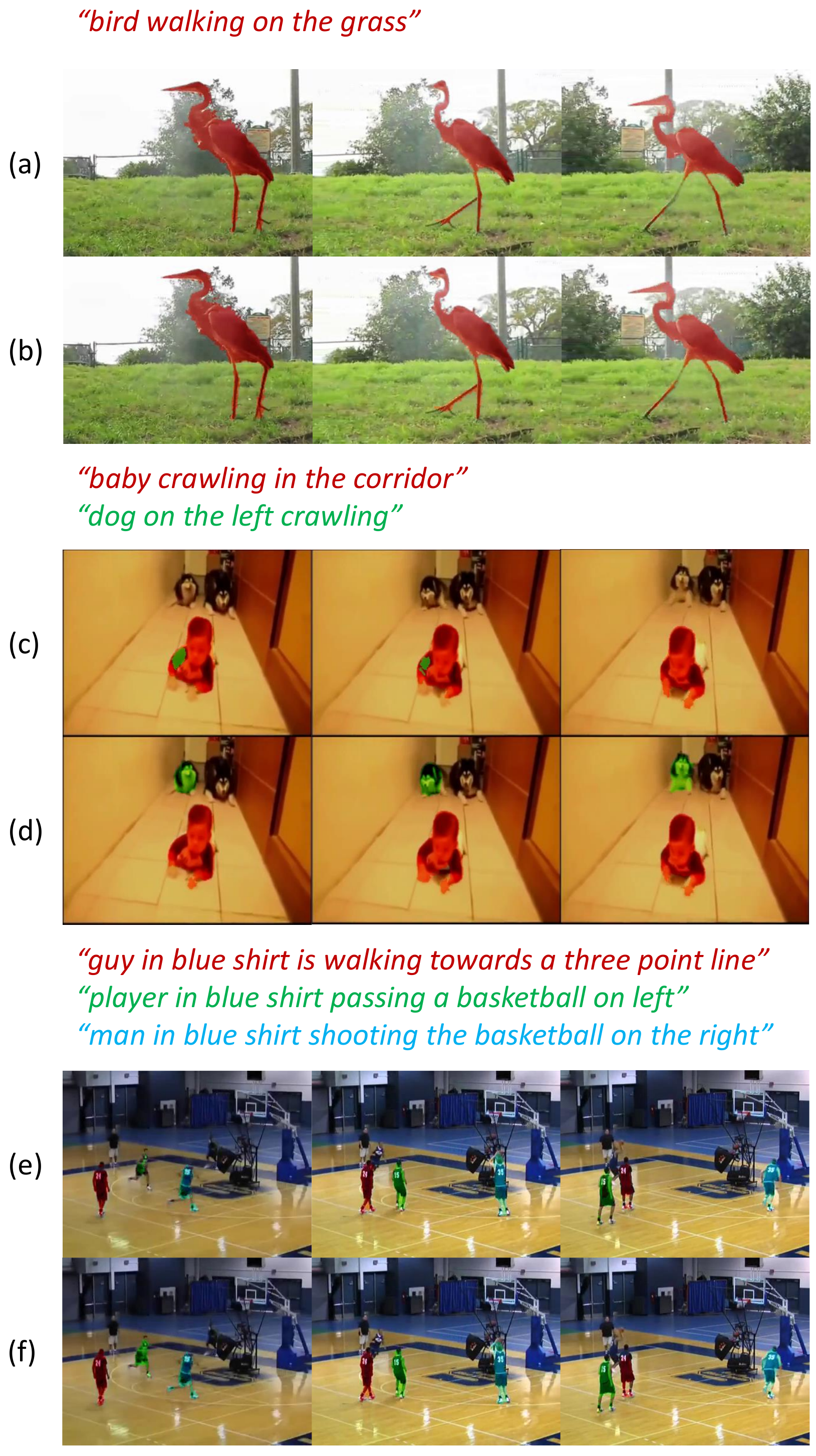}
    \end{center}
       \caption{Qualitative results of referring video segmentation. (a)(c)(e) Results of baseline model (row $1$ in Table~\ref{tab:cmpc_v}) (b)(d)(f) Results of our full video model (row $5$ in Table~\ref{tab:cmpc_v}). Colors of masks correspond to different expressions.}
    \label{fig:quality-v}
 \end{figure}
 
 \begin{figure*}[!htbp]
    \begin{center}
       \includegraphics[width=0.75\linewidth]{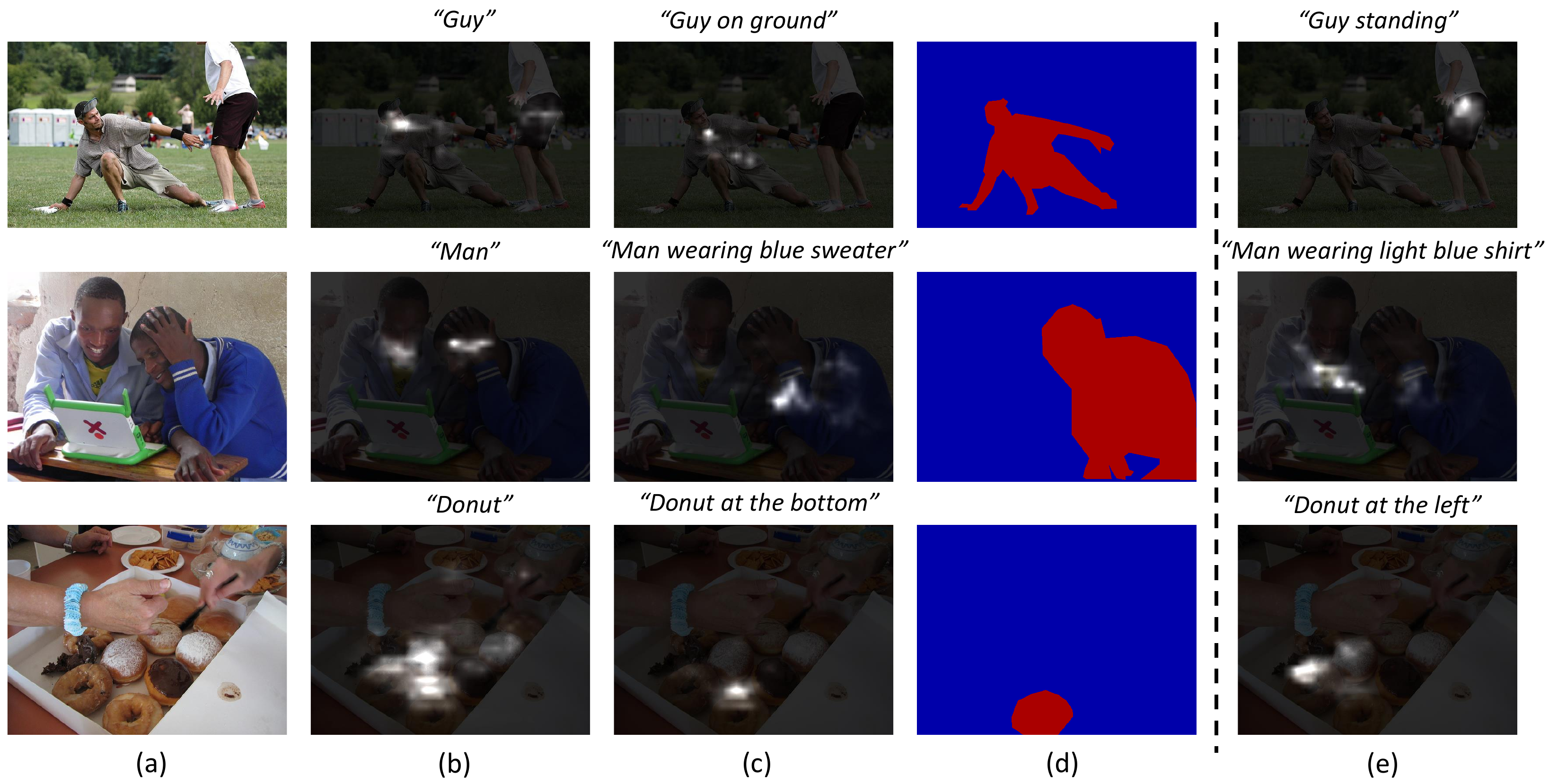}
    \end{center}
       \caption{Visualization of affinity maps between images and expressions in our CMPC-I module. (a) Original image. (b)(c) Affinity maps of only entity words and full expressions in the test samples. (d) Ground-truth. (e) Affinity maps of expressions manually modified by us.}
    \label{fig:attn_map}
 \end{figure*}

\subsubsection{Qualitative Results}
In Fig.~\ref{fig:quality-i}, we presents qualitative comparison between the multi-level baseline model (row $7$ in Table~\ref{tab:cmpc_i}) and our full model (row $13$ in Table~\ref{tab:cmpc_i}) for referring image segmentation. 
From the top-left example we can observe that the baseline model fails to make clear judgement between the two girls, while our model is able to distinguish the correct girl involving the relationship with the phone. 
Similar result is shown in the top-right example of Fig.~\ref{fig:quality-i}. 
As illustrated in the bottom row of Fig.~\ref{fig:quality-i}, attributes and location relationship can also be well handled by our model, demonstrating its effectiveness.

We also provide qualitative results of our full video model (row $5$ in Table~\ref{tab:cmpc_v}) and baseline model (row $1$ in Table~\ref{tab:cmpc_v}) in Fig.~\ref{fig:quality-v}. 
Colors of expressions correspond to masks of instances. 
As shown in Fig.~\ref{fig:quality-v} (c) and (d), our full video model well segments the baby and dog with coherent masks while the baseline model fails to distinguish different actors, indicating the effectiveness of our proposed modules.

\subsubsection{Visualization of Affinity Maps}
In Fig.~\ref{fig:attn_map}, we visualize the affinity maps between multimodal feature and the first word of the expression in our CMPC-I module. 
As shown in (b) and (c), our model can progressively produce more concentrated responses on the referent as the expression becomes more informative from only entity words to the full sentence. 
It should be noticed that when we manually modify the expression to refer to other entities in the image, our model can still comprehend the new expression and correctly identify the referent. 
For example, in the third row of Fig.~\ref{fig:attn_map}(e), when the expression changes from ``Donut at the bottom'' to ``Donut at the left'', high response area shifts from bottom donut to the left donut accordingly. 
It indicates that our model can adapt to new expressions flexibly.

\begin{figure}[htbp]
   \centering 
   \includegraphics[width=1.00\linewidth]{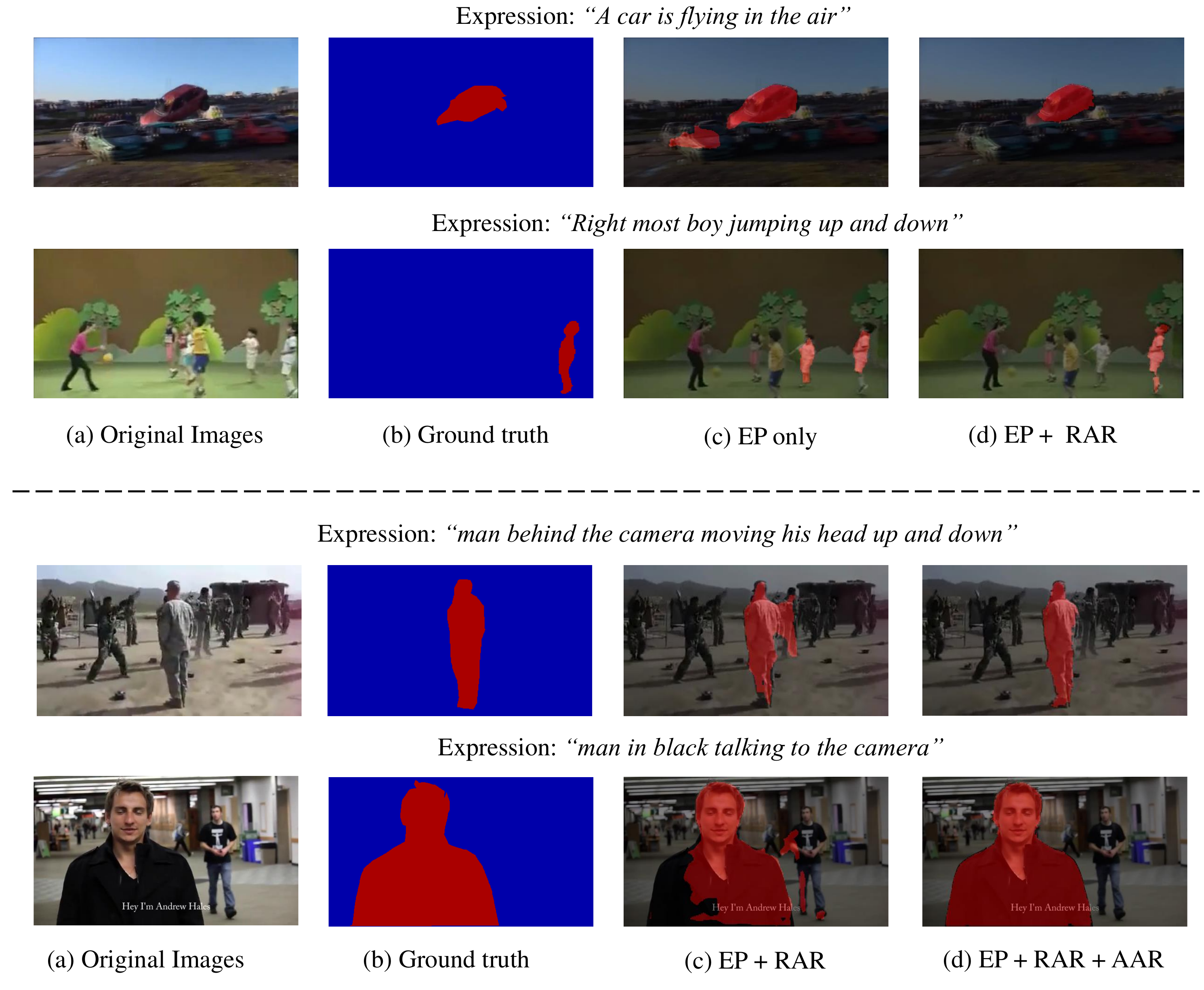}
   \caption{\textit{Top}: Segmentation results w/ and w/o relation-aware reasoning.  \textit{Bottom}: Segmentation results w/ and w/o action-aware reasoning.}
   \label{fig:aar_rar}
\end{figure}

\subsubsection{Effectiveness of Relation-Aware Reasoning and Action-Aware Reasoning}
To verify the effectiveness of relation-aware reasoning and action-aware reasoning, we present the comparison with and without them in Fig.~\ref{fig:aar_rar}.

First, we show the comparison of with relation-aware reasoning (EP + RAR) and without relation-aware reasoning (EP) in the top rows of Fig.~\ref{fig:aar_rar}. It is shown that comparing with EP, EP + RAR is able to discriminate the right referent from others. Taking the $2$nd row as an example, there are several boys in the images and EP only fails to recognize the right most boy. While EP + RAR makes the right prediction.

In addition, we also show the comparison of with action-aware reasoning (EP + RAR + AAR) and without action-aware reasoning (EP + RAR) in the bottom of Fig.~\ref{fig:aar_rar}. In the first row, EP + RAR fails to discriminate the man who is moving his head up and down from others who also stand behind camera. With AAR to recognize action in the video, EP + RAR + AAR is able to locate the correct man. The visualization results demonstrate the necessity of action reasoning.

\section{Conclusion and Future Work}

To address the referring segmentation problem for image and video, we propose a CMPC scheme which first perceives candidate entities which might be considered by the expression using entity and attribute words, then conduct graph-based reasoning with the aid of relational words and action words to further highlight the referent while suppressing others. 
We implement CMPC scheme as two modules, namely CMPC-I and CMPC-V for image and video inputs. 
We also propose a TGFE module which exploits textual information to selectively integrate multi-level features to refine the mask prediction. 
Our model consistently outperforms previous state-of-the-art methods on four referring image segmentation benchmarks and three referring video segmentation benchmarks, demonstrating its effectiveness. 
In the future, we plan to analyze the linguistic information more structurally and explore more compact graph formulation. 
Our code is available at \url{https://github.com/spyflying/CMPC-Refseg}.

\bibliographystyle{plain}
\bibliography{cmpc_trans}

\end{document}